\documentclass[letterpaper]{article} 
\usepackage[preprint]{aaai2027}  
\usepackage[hyphens]{url}  
\usepackage{graphicx} 
\usepackage{natbib}  
\usepackage{caption} 
\usepackage{booktabs}
\usepackage{multirow}
\usepackage{array}

\newcommand{\method}{\textsc{IAR}}
\newcommand{\inject}{\textsc{Inject}}
\newcommand{\align}{\textsc{Align}}
\newcommand{\recover}{\textsc{Recover}}
\newcommand{\na}{/}

\newcommand{\suppsec}[1]{Supplementary Material, Section~#1}

\title{Inject, Align, Recover: Staged Post-Training for Retrieval-Free Document Knowledge Internalization}
\author{
Qian Kou\textsuperscript{1}\equalcontrib\thanks{\raggedright Corresponding authors: Qian Kou (kouqian@baai.ac.cn) and Xiaofeng Shi (xfshi@baai.ac.cn).},\quad Xiaofeng Shi\textsuperscript{1}\equalcontrib\footnotemark[2],\quad Xiaosong Qiu\textsuperscript{1},\quad Hua Zhou\textsuperscript{1}\thanks{\raggedright Project leader.}
}
\affiliations{
\textsuperscript{1}Beijing Academy of Artificial Intelligence (BAAI)
}

\begin{document}
\maketitle
\begin{abstract}
Large language models often fail to answer questions about a bounded document collection when the source documents are not retrieved at inference time. We study this setting as document knowledge internalization: converting a fixed corpus into usable parametric knowledge for retrieval-free question answering. We propose \method{} (Inject, Align, and Recover), a three-stage post-training framework that separates structured document knowledge injection, QA behavior alignment, and general ability recovery. Unlike conventional continued pretraining, Inject converts source documents into continuation, rewrite, and instruction-conditioned reconstruction objectives. Align then adapts the injected model with answer-only QA supervision, while Recover merges the domain-adapted model with the base instruction model to recover general capabilities. Across Common Corpus (CC) and CCI, and across Llama, Phi, Qwen, and SmolLM model families, \method{} improves the domain-primary domain-general frontier for retrieval-free document internalization. In the main comparison, \method{} improves over Vanilla SFT on all four reported metrics in 7 of 8 dataset-model settings, with average gains of 3.6 percentage points in domain QA accuracy and 12.1 percentage points in mean general performance across IFEval, MMLU, and MSBench. Extended CC baselines show that LoRA and FAPM can win individual general metrics, but among methods that also reach leading or near-leading domain internalization, \method{} retains one of the strongest general profiles.
\end{abstract}

\section{Introduction}

Retrieval-augmented generation is the standard engineering answer to document-grounded question answering: retrieve passages from a corpus, append them to the prompt, and ask a model to answer with that evidence \citep{lewis2020retrieval}. This paper studies a different setting. In many deployments, retrieval may be unavailable, undesirable for latency or privacy reasons, or deliberately removed to test whether post-training has changed the model's parametric knowledge. We call this setting \emph{document knowledge internalization}. Given a bounded document collection and document-derived questions, the model must answer held-out questions without receiving the source documents at inference time. We evaluate this problem on Common Corpus (CC) \citep{commoncorpus} and CCI \citep{cci}, and use the abbreviations CC and CCI throughout the paper.

The simplest solution is supervised fine-tuning (SFT) on document-derived question-answer pairs. This approach gives the model the same input-output format that it will see at test time, but the learning signal is sparse: only the facts selected by the QA generator contribute to the loss. Continued pretraining (CPT) provides denser exposure to document text \citep{gururangan-etal-2020-dont}, but it does not directly teach the model how to answer questions. Both routes also face a general-capability trade-off: domain adaptation can improve target QA behavior while degrading instruction following or broad benchmark performance, a form of catastrophic forgetting \citep{kirkpatrick2017overcoming}. Document internalization therefore requires both domain acquisition and controlled recovery.

\begin{figure*}[t]
  \centering
  \includegraphics[width=0.98\textwidth]{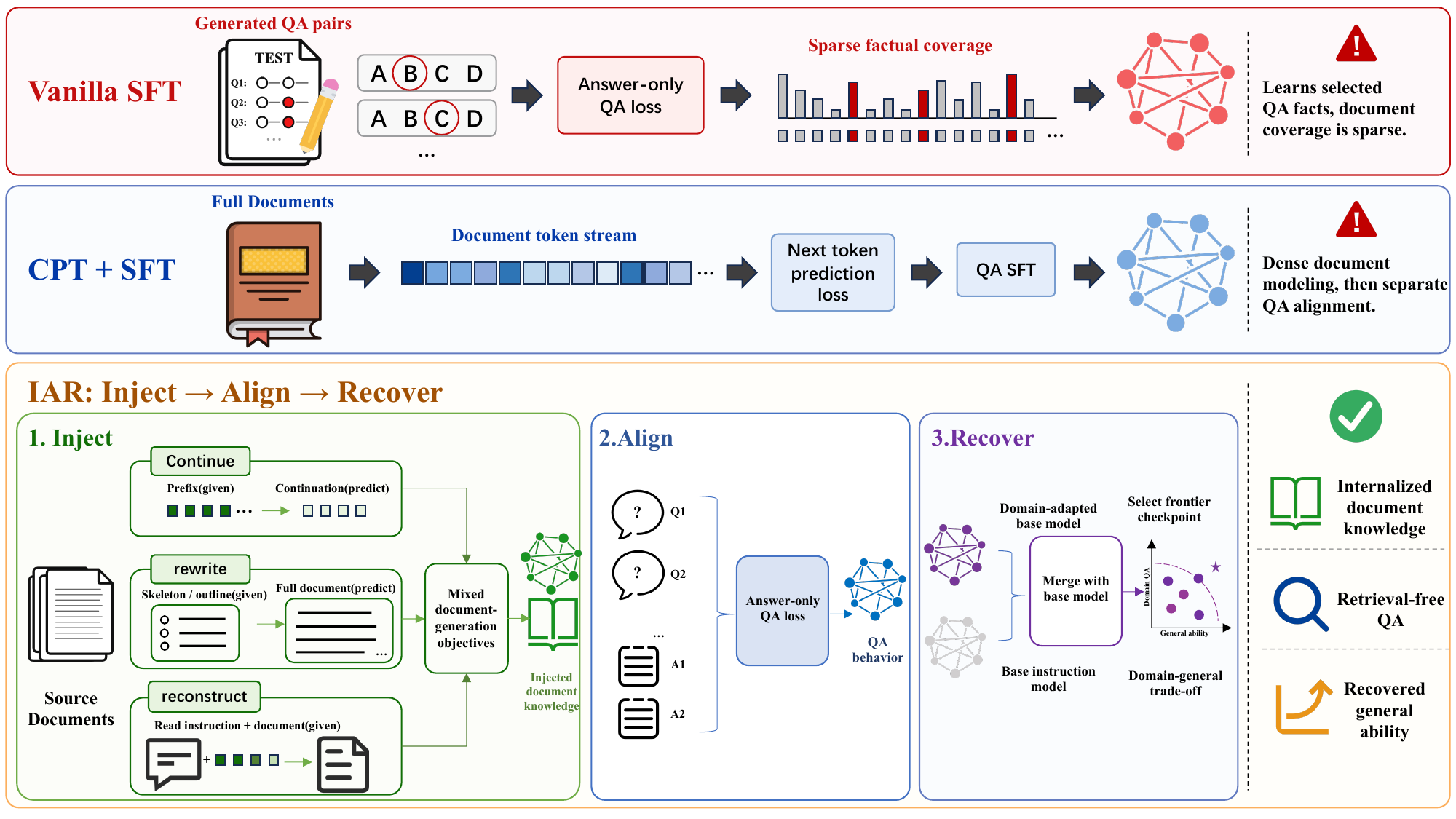}
  \caption{The overview of \method{}. Vanilla SFT learns from generated QA pairs and covers only the facts selected by those questions. CPT+SFT models full document token streams before a separate QA alignment stage. \method{} injects document knowledge through continuation, rewrite, and instruction-conditioned reconstruction objectives, aligns the injected model with answer-only QA supervision, and recovers general ability via post-hoc model merging. The final model is selected by balancing retrieval-free domain QA performance and general capability retention.}
  \label{fig:overview}
\end{figure*}

We propose \method{}, a three-stage framework that separates these functions. Unlike raw CPT, \method{} does not merely continue language modeling on domain text. \inject{} converts documents into instruction-conditioned supervised reconstruction tasks; \align{} maps the injected knowledge to a QA interface; and \recover{} performs post-hoc model merging between the domain-adapted checkpoint and the original instruction model, producing candidate checkpoints that trade domain accuracy against general capability. Figure~\ref{fig:overview} summarizes the contrast with Vanilla SFT and CPT+SFT.

The central empirical claim is that \method{} improves the operating-point frontier for retrieval-free document internalization. In multiple settings, \method{} improves domain accuracy over Vanilla SFT while restoring general capability that is often damaged by domain adaptation. Against BudgetMatch QA-only SFT, both CCI settings are all-metric wins for \method{}, with Qwen3-4B CCI showing the largest separation. More broadly, the experiments show that document exposure, answer alignment, data recipe, token budget, and recovery should be measured separately rather than collapsed into one fine-tuning comparison.

This paper makes four contributions:
\begin{itemize}
  \item We study retrieval-free QA over a fixed ingested corpus as a domain-general operating-point problem, measuring both domain accessibility and general capability.
  \item We present \method{}, a three-stage post-training framework that separates structured document exposure, answer-only QA alignment, and post-hoc general-capability recovery.
  \item We evaluate across CC and CCI and multiple model families, with direct SFT, LoRA, SDFT, Replay, FAPM, conventional Base-initialized continued-pretraining baselines, and available Instruct-initialized CPT diagnostics.
  \item We report both strong and boundary evidence: \method{} improves the domain-general frontier in most settings, while boundary cases clarify when initialization strength, data recipe, or operating-point selection matters most.
\end{itemize}

\section{Related Work}

\paragraph{Parametric knowledge and retrieval.}
Language models store factual associations in their parameters, but recall and updating remain unreliable \citep{petroni2019language,roberts2020how}. Retrieval-augmented systems externalize document memory through a retriever \citep{karpukhin2020dense,lewis2020retrieval}, while knowledge editing targets localized factual changes \citep{mitchell2022fast,meng2022locating,wang2024wise,cohen2024ripple}. We instead ask whether post-training makes a bounded corpus answerable without retrieval.

\paragraph{Document knowledge acquisition.}
The closest work changes the interface through which models acquire document knowledge. AdaptLLM converts raw corpora into reading-comprehension texts to couple domain content with task practice \citep{cheng2024adaptllm}. PIT reverses the conventional document-then-QA order by instruction-tuning on questions before continued document training \citep{jiang2024knowledgelearners}. SELF-TUNING augments unseen documents with self-supervised memorization, comprehension, and reflection tasks \citep{zhang2025selftuning}, while KiDG converts multiple documents into simulated dialogues for retrieval-free domain transfer \citep{wang2023kidg}. Complementary studies compare unsupervised fine-tuning with retrieval \citep{ovadia2024finetuning} and isolate how QA versus article-style supervision changes factual learning \citep{zhao2025stylefacts}.

These methods differ in corpus, model, training order, and evaluation contract, so their published scores are not protocol-aligned comparisons. \method{} does not claim a new family of reconstruction losses. Its increment is a controlled empirical decomposition: it separates structured document exposure from answer-only accessibility, then treats post-hoc capability recovery as an explicit third variable. We jointly evaluate the selected checkpoint on retrieval-free domain QA and general benchmarks across two corpora and multiple model families.

\paragraph{Forgetting and recovery.}
Task adaptation can damage general capability through catastrophic forgetting \citep{kirkpatrick2017overcoming}. FAPM mitigates this effect through post-hoc pruning \citep{huang2025fapm}, while model merging combines checkpoints or task vectors in weight space \citep{wortsman2022model,matena2022merging,ilharco2023editing,yadav2023ties,yu2023supermario}. \recover{} selects among existing SLERP \citep{shoemake1985animating}, task-arithmetic, TIES, and DARE operators rather than introducing a new merge algorithm.

\section{Methodology}
\label{sec:method}

\subsection{Task Definition}
\label{sec:problem}

Let $D=\{d_i\}_{i=1}^{N}$ be a target document collection. From $D$, we derive a training QA set $Q_{\mathrm{train}}$ and a held-out test QA set $Q_{\mathrm{test}}$. The training framework further splits $Q_{\mathrm{train}}$ into training and validation subsets for checkpoint monitoring and Recover-candidate selection. Given an instruction-tuned model $M_0$, the goal is to produce a model $M$ that answers questions in $Q_{\mathrm{test}}$ without seeing retrieved passages from $D$ at inference time. The primary domain metric is correctness-based QA accuracy. General capability is measured with IFEval \citep{zhou2023instruction}, MMLU \citep{hendrycks2021measuring}, and the public MSBench data derived from MSAgent-Bench \citep{li2023modelscope,msbench}. \suppsec{D} specifies the fixed 200-example MSBench subset and our evaluation protocol.

This setup differs from ordinary task SFT because the training questions expose only a subset of the document facts. It differs from CPT because the target behavior is instruction-following QA rather than document continuation. It differs from RAG because the model cannot defer memory to a retriever at test time.

\subsection{Why Three Stages?}

The three stages are motivated by three testable hypotheses concerning document exposure, QA accessibility, and capability retention. First, QA-only supervision may expose too little of the corpus. Second, document-level training may not make the acquired information accessible under a question-answering interface. Third, target-domain adaptation may damage general instruction-following ability. Collapsing these hypotheses into one fine-tuning recipe makes it hard to identify which intervention explains a gain or failure.

\method{} is therefore designed less as a single fixed recipe than as an experimental decomposition. \inject{} asks whether denser document-level supervision helps before QA alignment. \align{} asks whether the exposed information can be made accessible through question answering. \recover{} asks whether the adapted model can be moved back toward the original instruction model without losing the domain behavior that was acquired. This decomposition is important for interpretation: a negative result in one stage does not invalidate the others, and a strong SFT baseline can be understood as a competing way to spend the same adaptation budget.

\subsection{A Unified Objective View}

Let $\mathcal{D}_{\mathrm{doc}}=\{d_i\}$ denote the target documents and $\mathcal{D}_{\mathrm{QA}}=\{(q_i,a_i)\}$ denote document-derived QA pairs. Each Inject objective $m$ constructs a recipe dataset $\mathcal{D}_m=\{(u,y)\}$, where $u$ is an instruction with an optional document prefix or compressed representation and $y$ is the supervised document target. Let $n_m$ be its realized row count after mixture sampling and length filtering, and let $\pi_m=n_m/\sum_k n_k$. The Inject objective is
\begin{equation}
\begin{array}{rl}
\mathcal{L}_{\mathrm{inj}}
 &= \displaystyle\sum_{m\in\mathcal{M}}\pi_m
    \mathrm{E}_{(u,y)\sim\mathcal{D}_m}
    [\ell_{\theta}(u,y)],\\[2pt]
\ell_{\theta}(u,y)
 &= \displaystyle-\frac{1}{|y|}\sum_{t=1}^{|y|}
    \log p_{\theta}(y_t\mid u,y_{<t}).
\end{array}
\end{equation}
Thus $\pi_m$ is the empirical sampling share, not a free loss coefficient. The system and user prompt tokens are masked; loss is applied only to the assistant target $y$. This differs operationally from raw continued pretraining:
\begin{equation}
  \mathcal{L}_{\mathrm{CPT}}
  =-\frac{1}{T}\sum_{t=1}^{T}\log p_{\theta}(x_t\mid x_{<t}).
\end{equation}
CPT models the document stream directly and has no explicit prompt/target boundary or QA interface. In our recipe labels, 1:0:0 denotes the reconstruction-only \inject{} recipe with a reading prompt; it is not the raw CPT baseline. \suppsec{B} defines every objective and its masking rule.

\subsection{Stage 1: Inject}

The \inject{} stage uses three document-generation objectives and their mixtures. \emph{Continuation} predicts a document suffix from an instruction-conditioned prefix. \emph{Rewrite} reconstructs the cleaned document from a generated summary, outline, or knowledge skeleton. \emph{Instruction-formatted reconstruction} predicts the cleaned document from a short generic reading instruction. These objectives provide denser supervised document targets than QA-only training without using raw-stream CPT loss. \suppsec{B} gives the exact inputs, targets, masks, mixture notation, and prompt schemas; \suppsec{C} gives training hyperparameters and realized row counts.

\subsection{Stage 2: Align}

The \align{} stage fine-tunes the injected model on domain QA pairs. For a question $x$ and answer $y$, we use answer-only supervised fine-tuning:
\begin{equation}
  \mathcal{L}_{\mathrm{align}}
  =-\frac{1}{|a|}\sum_{t=1}^{|a|}
  \log p_{\theta}(a_t\mid q,a_{<t}).
\end{equation}
Vanilla SFT optimizes this loss from the original instruction model $\theta_0$, while \method{} optimizes it from the injected checkpoint $\theta_I$:
\begin{equation}
  \theta_{\mathrm{SFT}}=\mathrm{Align}(\theta_0),
  \qquad
  \theta_{\mathrm{IA}}=\mathrm{Align}(\theta_I).
\end{equation}
We use \textbf{IA} to denote this pre-recovery Inject+Align checkpoint. BudgetMatch keeps the same QA-only objective but uses setting-specific epoch counts matched to the token budget of the two-stage IA pipeline, following the broader observation that token count and compute allocation can change adaptation behavior \citep{kaplan2020scaling,hoffmann2022training}. \suppsec{C} reports the calculation and realized per-setting token counts.

\subsection{Stage 3: Recover}

The \recover{} stage starts from the original instruction model $M_0$ and a domain-adapted checkpoint $M_{\mathrm{IA}}$. Its purpose is to mitigate the catastrophic-forgetting side of document adaptation: a checkpoint can internalize the target corpus while losing instruction-following or broad benchmark capability. The simplest view is task-vector interpolation,
\begin{equation}
  \Delta=\theta_{\mathrm{IA}}-\theta_0,\qquad
  \theta_R=\theta_0+\lambda\Delta.
\end{equation}
More generally, \recover{} evaluates post-hoc merge operators $\theta_R=\mathrm{Merge}(\theta_0,\theta_{\mathrm{IA}})$ from four families: SLERP, task arithmetic, TIES, and DARE. The selected checkpoint is not necessarily the highest-domain checkpoint. We use a domain-primary frontier criterion on the validation split: domain accuracy is the main objective, while IFEval, MMLU, and MSBench are guardrails for deployability. The held-out test set is used only after the merge candidate is fixed. \suppsec{E} reports the selection rule and selected recovery settings, and \suppsec{G} gives the full pre-recovery recipe grids.

\paragraph{Recover selection.}
For a candidate $c$, let $D(c)$ be validation-domain accuracy and let $G(c)$ be the mean of validation IFEval, MMLU, and MSBench. With Vanilla SFT denoted by $v$ and a fixed tolerance $\tau=1.0$ percentage point, we first retain candidates satisfying $D(c)\geq D(v)-\tau$. We then require $G(c)\geq G(v)$ and require at least two of the three general metrics to be no more than $\tau$ below their Vanilla SFT values. Among non-dominated candidates in the $(D,G)$ plane, domain accuracy is the primary key; candidates within $\tau$ of the best remaining domain score form one domain tier, within which we choose the largest $G(c)$. Remaining ties use the largest minimum general-metric improvement and then the smaller within-family merge hyperparameter. This fixed validation rule is applied before any held-out test evaluation. Thus the reported test frontiers diagnose the selected operating points but do not select them.

\section{Experiments}
\label{sec:experiments}

We ask four questions: \textbf{RQ1}, whether \method{} improves retrieval-free domain-general operating points over direct SFT and conventional CPT+SFT; \textbf{RQ2}, whether extended QA-only training explains the gains; \textbf{RQ3}, whether Inject+Align improves domain internalization before Recover; and \textbf{RQ4}, whether the recovery pattern persists across larger Qwen3 models on CC. The CC-only SDFT, LoRA, Replay, and FAPM stress test is grouped under RQ1, while Qwen scaling remains separate from cross-family comparisons.

For every recoverable IA checkpoint, the Recover experiment evaluates a fixed 12-candidate grid: SLERP with $t\in\{0.2,0.3,0.4\}$, task arithmetic with $w\in\{0.3,0.5,0.7\}$, TIES with $d\in\{0.3,0.5,0.7\}$, and DARE with $d_r\in\{0.1,0.3,0.5\}$. This grid is fixed before validation-based selection; the selected operating point is then reported on the held-out test set. \suppsec{E} gives the candidate-grid summary and the selected operating points.

\subsection{Datasets and Models}

CC contains 14,258 training and 750 test QA pairs derived from Common Corpus \citep{commoncorpus}; CCI contains 10,926 training and 575 test pairs derived from CCI \citep{cci}. Test inputs contain only the question. Both datasets cover Llama-3.2-3B, Phi-4-mini, Qwen3-4B, and SmolLM3-3B, while Qwen3-8B/14B/32B are reported as CC scaling ablations. \suppsec{A} gives the dataset contract, and \suppsec{B} gives the prompt templates.

We keep the main comparison focused on rows that support the central domain-general claim. CPT+SFT diagnostics are included as dense-document baselines but are not used to define the \method{} operating-point frontier.

\subsection{Baselines}

We compare with the original instruction model, Vanilla SFT, BudgetMatch, SDFT, LoRA, Replay, Base-initialized CPT+SFT, and FAPM. These controls probe QA supervision, token budget, supervised-data recipe, parameter-efficient adaptation, anti-forgetting replay, raw-document modeling, and pruning-based recovery, respectively. Context-conditioned SFT is not a reported baseline because evaluation supplies no retrieval context.

We report CPT+SFT from the corresponding released Base checkpoint as the conventional dense-document baseline. This starting point differs from the Instruct initialization used by \method{} and therefore does not, by itself, isolate the effect of the Inject objective. \suppsec{G} reports the available Instruct-initialized CPT+SFT diagnostics for Llama on CC and CCI and for Phi on CC. Phi-4-mini has no corresponding non-instruction Base release and is therefore unavailable in the main CPT+SFT comparison.

\subsection{Evaluation}

Domain QA is evaluated with an adaptive LLM panel. Two judges score each answer; a third judge is queried only when the first two scores differ, and the final ordinal score is their median. Per-judge outputs are stored before aggregation. The archived panel configurations use \texttt{gpt-oss-120b} \citep{gptoss120b}, \texttt{MiniMax-M2.5} \citep{minimaxm25}, and DeepSeek-V3 variants \citep{deepseekv32}. Across 242,255 evaluated model-answer instances, the audit finds first-two exact agreement of .707, binary agreement of .848 after collapsing scores at $\geq .5$, binary Cohen's $\kappa=.691$, and a third-judge trigger rate of .297. LLM-as-judge evaluation is useful but imperfect, so we treat it as an auditable instrument rather than a gold label source \citep{zheng2023judging}. We report domain accuracy as the fraction of correct or partially correct answers; IFEval, MMLU, and MSBench measure general ability. Per-result-file bootstrap intervals use 2,000 resamples and quantify evaluation-sample uncertainty. \suppsec{C} records training runs, while \suppsec{D} gives decoding settings and the full reliability protocol.

\paragraph{Inference controls.}
Domain QA uses one generation pass per checkpoint with temperature .7, \texttt{top\_p=.95}, repetition penalty 1.1, and a 2,048-token limit. The reported non-thinking IFEval, MMLU, and MSBench runs use greedy decoding with token limits of 1,024, 10, and 1,024, respectively; MSBench judge calls use temperature .1. Domain generation and judge APIs receive no explicit sampler seed. We therefore treat fixed training and data-order seeds as insufficient to establish repeated-run robustness and preserve raw judge votes for audit.

\section{Results}
\label{sec:results}

\subsection{RQ1: Main Domain-General Operating Points}

Table~\ref{tab:main} reports the main comparison used for the central claim. The table includes the original instruction model, Vanilla SFT, conventional Base-initialized CPT+SFT when available, and \method{}. BudgetMatch is reported separately because it tests token budget rather than ordinary recipe choice. \suppsec{G} reports the CC-only extended-baseline stress test: LoRA has CC coverage only, and FAPM is a pruning-style recovery baseline whose intervention differs from the \method{} rows.

\begin{table*}[t]
\centering
\footnotesize
\setlength{\tabcolsep}{2pt}
\begin{tabular*}{\textwidth}{@{\extracolsep{\fill}}llrrrr|rrrr@{}}
\toprule
\multirow{2}{*}{Model} & \multirow{2}{*}{Method} & \multicolumn{4}{c|}{CC} & \multicolumn{4}{c}{CCI} \\
\cmidrule(lr){3-6}\cmidrule(lr){7-10}
 & & Dom. & IFEval & MMLU & MSB. & Dom. & IFEval & MMLU & MSB. \\
\midrule
\multirow{4}{*}{Llama-3.2-3B} & Base Instruct & 11.2 & \textbf{77.1} & \textbf{50.8} & \textbf{54.5} & 29.2 & \textbf{77.1} & \textbf{50.8} & \textbf{54.5} \\
 & Vanilla SFT & 35.5 & 54.2 & 11.2 & 21.5 & 53.0 & 61.2 & 22.5 & 31.5 \\
 & CPT+SFT & \textbf{38.3} & 26.4 & 3.7 & 13.5 & \underline{53.7} & 24.2 & 4.3 & 17.5 \\
 & \method{} & \underline{36.5} & \underline{60.2} & \underline{35.0} & \underline{30.5} & \textbf{55.3} & \underline{61.3} & \underline{33.2} & \underline{36.5} \\
\addlinespace
\multirow{4}{*}{Phi-4-mini} & Base Instruct & 13.1 & \textbf{77.2} & \textbf{61.0} & \textbf{62.0} & 27.3 & \textbf{77.2} & \textbf{61.0} & \textbf{62.0} \\
 & Vanilla SFT & \underline{24.4} & 47.8 & 51.0 & 32.5 & \textbf{40.2} & 47.8 & \underline{53.8} & 31.5 \\
 & CPT+SFT & \na{} & \na{} & \na{} & \na{} & \na{} & \na{} & \na{} & \na{} \\
 & \method{} & \textbf{34.1} & \underline{49.0} & \underline{57.0} & \underline{43.0} & \underline{39.7} & \underline{51.6} & 50.2 & \underline{44.0} \\
\addlinespace
\multirow{4}{*}{Qwen3-4B} & Base Instruct & 34.3 & \textbf{84.8} & \textbf{65.8} & \textbf{77.0} & 70.6 & \textbf{84.8} & \textbf{65.8} & \textbf{77.0} \\
 & Vanilla SFT & 42.4 & 51.1 & 8.8 & 51.0 & \underline{75.1} & 45.6 & 26.3 & 49.5 \\
 & CPT+SFT & \underline{49.6} & 31.8 & 18.8 & 60.5 & 69.0 & 29.6 & 12.8 & 58.0 \\
 & \method{} & \textbf{50.5} & \underline{59.8} & \underline{19.5} & \underline{63.0} & \textbf{76.3} & \underline{76.1} & \underline{64.5} & \underline{70.0} \\
\addlinespace
\multirow{4}{*}{SmolLM3-3B} & Base Instruct & 15.3 & \textbf{77.1} & \textbf{47.7} & \textbf{63.0} & 34.1 & \textbf{77.1} & \textbf{47.7} & \textbf{63.0} \\
 & Vanilla SFT & 32.1 & 35.6 & 10.5 & 25.0 & \underline{52.3} & 41.7 & 16.7 & 31.5 \\
 & CPT+SFT & \underline{37.1} & 26.9 & 4.8 & 26.0 & 48.9 & 24.1 & 4.2 & 19.5 \\
 & \method{} & \textbf{37.5} & \underline{40.3} & \underline{25.7} & \underline{29.0} & \textbf{53.9} & \underline{57.4} & \underline{46.8} & \underline{47.0} \\
\bottomrule
\end{tabular*}
\caption{Main comparison for RQ1. Scores are reported as percentages. Within each model block and dataset block, \textbf{bold} denotes the best result and \underline{underlining} denotes the second-best result for each metric. CPT+SFT starts from the corresponding released Base checkpoint; ``/'' indicates that no such release is available, as for Phi-4-mini.}
\label{tab:main}
\end{table*}

The main pattern is a domain-general frontier rather than uniform dominance. On CC, \method{} exceeds Vanilla SFT on domain accuracy and all three general metrics for all four model families; on CCI, this holds for Llama, Qwen3-4B, and SmolLM3-3B. Vanilla SFT often trades broad capability for domain accuracy, whereas \method{} recovers part of that capability while retaining domain gains. Qwen3-4B CC is the clearest example, reaching 50.5\% versus 42.4\% domain accuracy while improving all three general metrics. Llama and SmolLM show smaller domain gains, but their general recovery remains important because a retrieval-free internalization model that cannot follow ordinary instructions is not deployable.

CCI exposes two boundary cases. Qwen3-4B starts from an unusually high 70.6\% base domain score, leaving less headroom for adaptation; nevertheless, \method{} improves over Vanilla SFT on all four reported metrics. \suppsec{F} analyzes this setting and gives the source-text BPB diagnostic. For Phi, \method{} improves IFEval and MSBench while slightly reducing domain accuracy and MMLU relative to Vanilla SFT. Keeping this row in the main table makes clear that recovery can improve part of the general profile without producing a uniformly dominant point, so deployment preference still matters.

\paragraph{CC-only extended baseline stress test.}

The CC-only extended comparison stress-tests RQ1 with SDFT, LoRA, Replay, and FAPM \citep{huang2025fapm}. \suppsec{G} reports every baseline row and the Instruct-initialized CPT diagnostics, while conventional Base-initialized CPT+SFT remains in Table~\ref{tab:main}.

SDFT is a strong supervised-data baseline, especially for Llama and Phi, and Replay is competitive on several Qwen3-4B general metrics. LoRA and FAPM are also strong general-retention baselines: they often preserve or recover stronger individual general metrics than \method{}, but usually trail on domain internalization. \method{} is domain-best for Phi, Qwen3-4B, and SmolLM3-3B and second only to SDFT for Llama. Its advantage should therefore be read as a strong domain-primary operating point, not as uniform dominance on every general metric.

\subsection{RQ2: Token-Budget Matched QA-only SFT}

A central confound is whether \method{} benefits simply from more training tokens. Table~\ref{tab:budget} compares Vanilla SFT, BudgetMatch, and \method{}. BudgetMatch uses 14, 17, 21, and 11 QA-only epochs for CC Llama, CC Qwen3-4B, CCI Llama, and CCI Qwen3-4B, respectively, to match the Inject+Align token budget reported in \suppsec{C}. This control is strong but corpus-dependent: it raises domain accuracy over Vanilla SFT by 4.9 and 4.4 points on CC, but leaves CCI Llama effectively unchanged and lowers CCI Qwen3-4B by 2.9 points. Relative to BudgetMatch, \method{} has higher domain accuracy in three of four settings and a higher mean general score in all four.

\begin{table*}[t]
\centering
\footnotesize
\setlength{\tabcolsep}{2pt}
\begin{tabular*}{\textwidth}{@{\extracolsep{\fill}}llrrrr|rrrr@{}}
\toprule
\multirow{2}{*}{Model} & \multirow{2}{*}{Method} & \multicolumn{4}{c|}{CC} & \multicolumn{4}{c}{CCI} \\
\cmidrule(lr){3-6}\cmidrule(lr){7-10}
 & & Dom. & IFEval & MMLU & MSB. & Dom. & IFEval & MMLU & MSB. \\
\midrule
\multirow{3}{*}{Llama-3.2-3B} & Vanilla SFT & 35.5 & \underline{54.2} & 11.2 & 21.5 & \underline{53.0} & \underline{61.2} & \underline{22.5} & \underline{31.5} \\
 & BudgetMatch & \textbf{40.4} & 53.4 & \underline{17.3} & \underline{22.0} & \underline{53.0} & 45.3 & \underline{22.5} & 24.0 \\
 & \method{} & \underline{36.5} & \textbf{60.2} & \textbf{35.0} & \textbf{30.5} & \textbf{55.3} & \textbf{61.3} & \textbf{33.2} & \textbf{36.5} \\
\addlinespace
\multirow{3}{*}{Qwen3-4B} & Vanilla SFT & 42.4 & \underline{51.1} & 8.8 & \underline{51.0} & \underline{75.1} & 45.6 & 26.3 & 49.5 \\
 & BudgetMatch & \underline{46.8} & 49.4 & \textbf{31.8} & 45.0 & 72.2 & \underline{56.1} & \underline{29.5} & \underline{52.5} \\
 & \method{} & \textbf{50.5} & \textbf{59.8} & \underline{19.5} & \textbf{63.0} & \textbf{76.3} & \textbf{76.1} & \textbf{64.5} & \textbf{70.0} \\
\bottomrule
\end{tabular*}
\caption{Token-budget ablation for RQ2 (higher is better; scores are percentages). BudgetMatch uses setting-specific QA-only epochs matched to the Inject+Align budget. \method{} improves both domain accuracy and mean general performance in three settings; CC Llama is the remaining domain--general trade-off. \suppsec{C} gives the full accounting.}
\label{tab:budget}
\end{table*}

\begin{figure}[!b]
  \centering
  \includegraphics[width=\columnwidth]{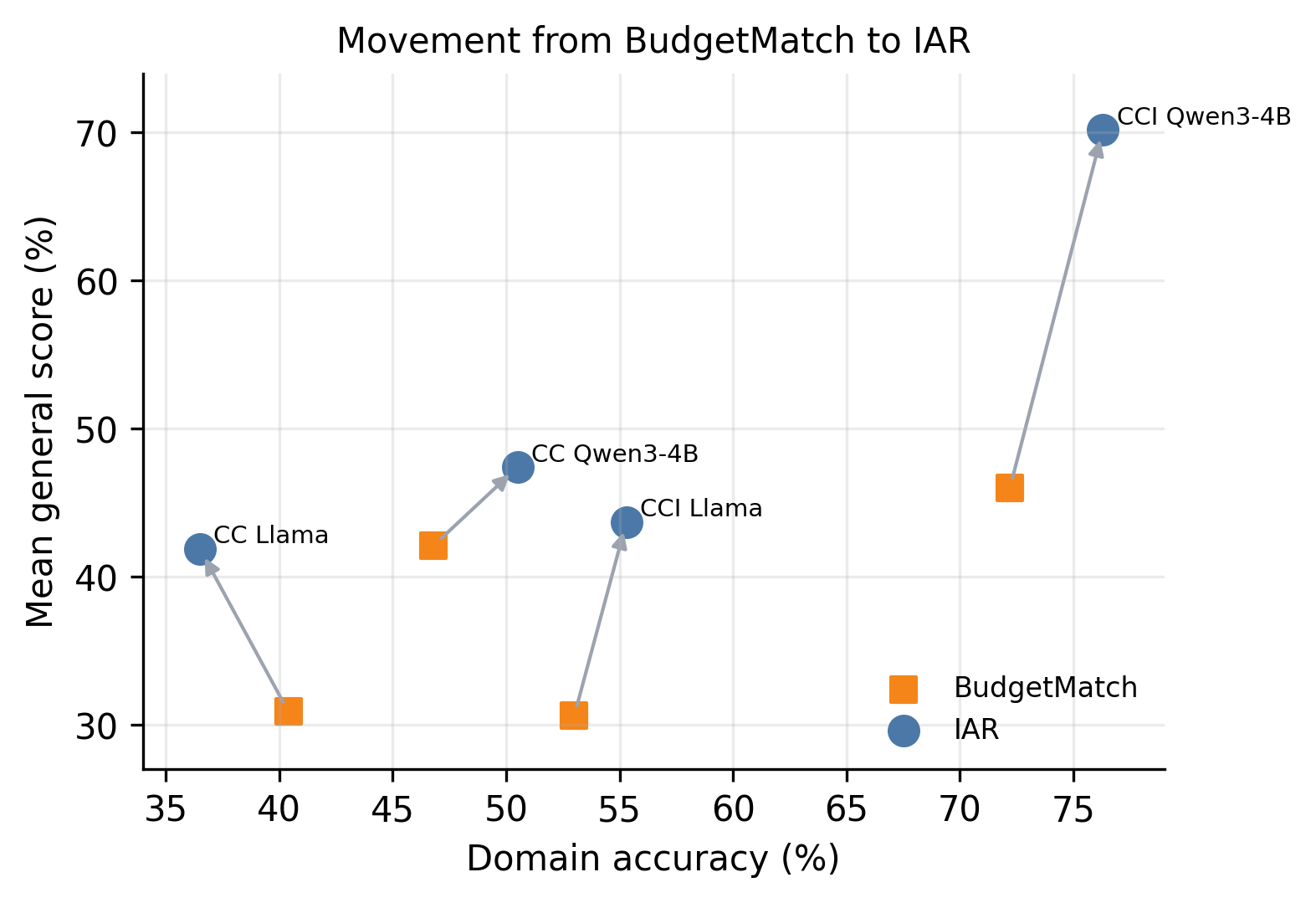}
  \caption{BudgetMatch-to-\method{} movement. Right is higher domain accuracy; up is a higher mean over IFEval, MMLU, and MSBench. \method{} moves up and right in three of four settings; for CC Llama, it trades 3.9 domain points for an 11.0-point gain in mean general performance.}
  \label{fig:recovery-frontier}
\end{figure}

BudgetMatch sharpens the token-budget diagnosis. Repeated QA-only training is competitive on CC, where it improves both domain accuracy and mean general performance over short Vanilla SFT, but this effect does not transfer uniformly to CCI. Figure~\ref{fig:recovery-frontier} shows that \method{} dominates BudgetMatch in the domain--mean-general projection for CC Qwen3-4B and both CCI settings; CC Llama remains a genuine trade-off. At the metric level, \method{} wins 14 of 16 comparisons, with the exceptions being CC Llama domain accuracy and CC Qwen3-4B MMLU. Thus token allocation explains part of the CC adaptation gain, but not the stronger operating points produced by staged document exposure and recovery.

\subsection{RQ3: Pre-Recovery Domain Internalization}

\begin{figure}[t]
  \centering
  \includegraphics[width=\columnwidth]{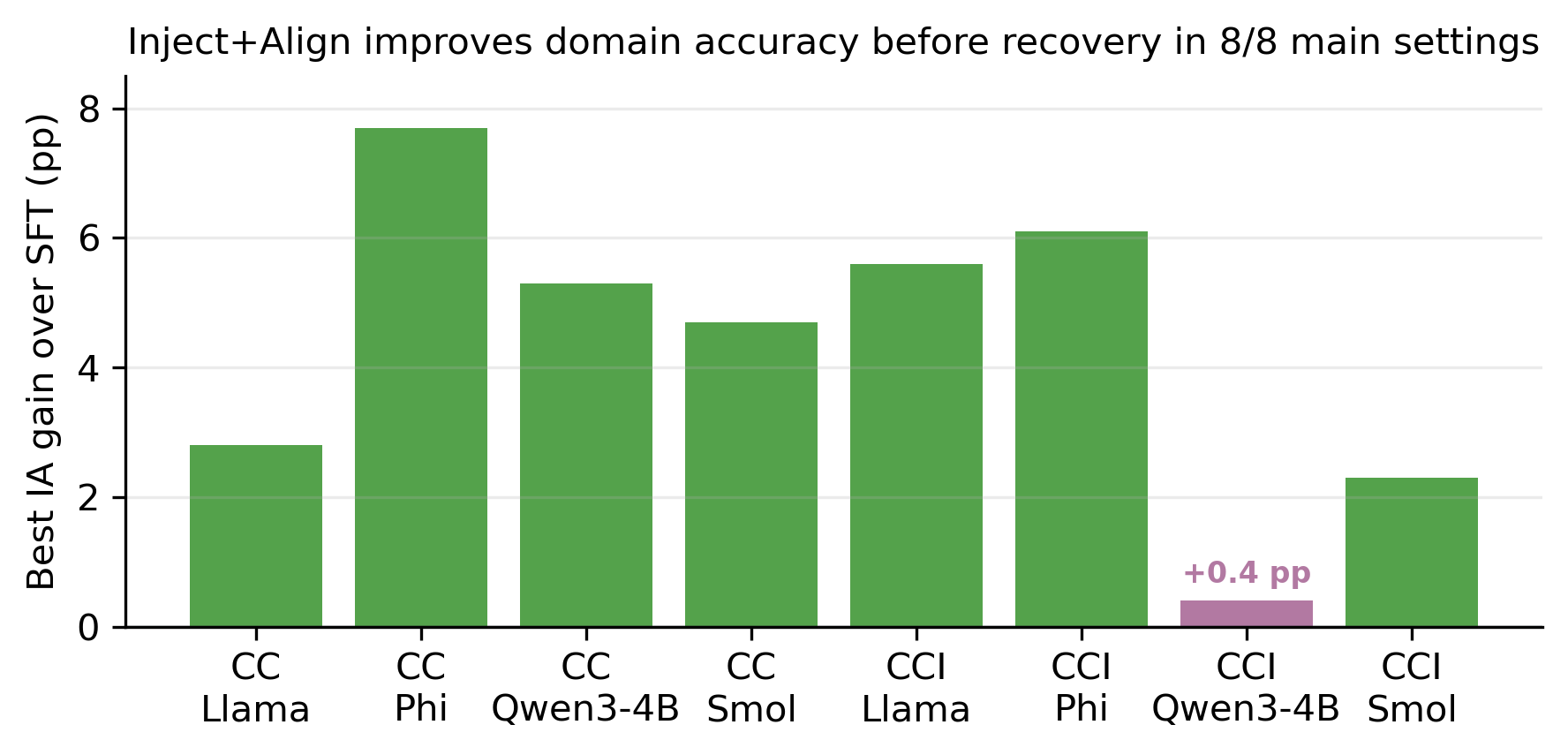}
  \caption{Pre-recovery domain gains from Inject+Align. CCI Qwen3-4B's $+0.4$ pp is the high-base-prior boundary case.}
  \label{fig:ia-gain}
\end{figure}

Recover should not hide the domain signal contributed by the first two stages. Figure~\ref{fig:ia-gain} shows that the best pre-recovery Inject+Align checkpoint improves domain accuracy over Vanilla SFT in all eight settings. Gains are largest for Phi and Llama and smallest for CCI Qwen3-4B, whose Vanilla baseline is already high.

No Inject recipe is uniformly best: Mixed 1:1:2 is strong for Llama and Phi, Qwen3-4B CC favors Mixed 1:1:1, and Qwen3-4B CCI slightly favors reconstruction-only 1:0:0. Thus document-level supervision helps, but the useful mixture remains model- and corpus-dependent. \suppsec{G} reports the exact scores and full recipe grid.

\paragraph{Stage-level effect.}
The pre-recovery gains are not driven by one model family. Relative to Vanilla SFT, Best IA improves domain accuracy by 2.8, 7.7, 5.3, and 4.7 points on CC for Llama, Phi, Qwen3-4B, and SmolLM, respectively; the corresponding CCI gains are 5.6, 6.1, 0.4, and 2.3 points. The 0.4-point Qwen3-4B CCI result is the only near-tie and coincides with the unusually strong 70.6\% initial checkpoint analyzed in \suppsec{F}. Therefore, RQ3 supports the narrower mechanism claim needed by the framework: structured document exposure contributes a measurable domain signal before any weight-space recovery is applied. It does not support a universal best Inject recipe.

\subsection{RQ4: Qwen Scaling on CC}

\begin{table*}[t]
\centering
\small
\setlength{\tabcolsep}{5pt}
\begin{tabular}{llrrrr}
\toprule
Model & Method & Domain (\%) & IFEval (\%) & MMLU (\%) & MSBench (\%) \\
\midrule
\multirow{4}{*}{Qwen3-8B} & Base Instruct & 38.5 & \textbf{87.6} & \textbf{65.3} & \textbf{82.5} \\
 & Vanilla SFT & 48.7 & 56.4 & 14.0 & 52.5 \\
 & Best IA & \textbf{57.5} & 50.6 & 18.5 & 48.5 \\
 & \method{} (TIES $d=0.3$) & \underline{56.8} & \underline{62.2} & \underline{26.7} & \underline{73.5} \\
\midrule
\multirow{4}{*}{Qwen3-14B} & Base Instruct & 40.4 & \textbf{90.0} & \textbf{72.5} & \textbf{81.5} \\
 & Vanilla SFT & 54.8 & 62.9 & 54.5 & 57.0 \\
 & Best IA & \textbf{60.5} & 53.5 & 40.3 & 42.0 \\
 & \method{} (TIES $d=0.3$) & \underline{59.6} & \underline{67.5} & \underline{67.2} & \underline{73.5} \\
\midrule
\multirow{4}{*}{Qwen3-32B} & Base Instruct & 47.2 & \textbf{87.5} & \textbf{74.8} & \textbf{84.5} \\
 & Vanilla SFT & 56.4 & 58.5 & 44.0 & 56.5 \\
 & Best IA & \textbf{63.9} & 53.0 & 63.0 & 44.5 \\
 & \method{} (TIES $d=0.3$) & \underline{62.8} & \underline{67.0} & \underline{74.5} & \underline{72.5} \\
\bottomrule
\end{tabular}
\caption{Complete Qwen3 scaling ablation on CC. Bold and underline mark the best and second-best result within each model block. Across 8B/14B/32B, \method{} stays within 1.1 points of Best IA domain accuracy while recovering 14.9--24.1 points in mean general performance.}
\label{tab:scaling}
\end{table*}

Table~\ref{tab:scaling} separates scale effects from cross-family comparisons. Across Qwen3-8B/14B/32B, the selected TIES $d=0.3$ checkpoint stays within 1.1 points of Best IA domain accuracy while raising mean general performance by 14.9--24.1 points. The claim is limited to this repeated CC pattern.

\paragraph{Per-benchmark recovery.}
The mean gains reflect improvements on every general benchmark, not compensation by a single metric. Relative to Best IA, \method{} raises IFEval/MMLU/MSBench by 11.6/8.2/25.0 points at 8B, 14.0/26.9/31.5 at 14B, and 14.0/11.5/28.0 at 32B. At the same time, its domain score decreases by only 0.7, 0.9, and 1.1 points, respectively. The repeated pattern is therefore stable across these three sizes: Recover trades a small amount of the maximum pre-recovery domain score for a much larger restoration of broad capability.

\paragraph{Adaptation and recovery remain separable.}
Scaling does not remove the contribution of Inject+Align. Relative to Vanilla SFT, Best IA adds 8.8, 5.7, and 7.5 domain points at 8B, 14B, and 32B; after Recover, \method{} still retains gains of 8.1, 4.8, and 6.4 points. Thus the larger-model result is not explained by merging an otherwise unchanged instruction checkpoint. Inject+Align first establishes a stronger domain checkpoint, and Recover then moves that checkpoint toward a more usable domain--general operating point. The magnitude of recovery varies by benchmark and size, but the two-stage empirical signature remains visible in every scaling block.

\paragraph{What does not scale away.}
Recovery remains partial rather than complete. The selected checkpoints retain 15.6--19.2 points more domain accuracy than the original instruction models, but they do not uniformly return to the original models' general scores. Moreover, all three rows use CC and the same TIES density, so RQ4 demonstrates a repeated within-family operating-point pattern rather than a scaling law or cross-corpus guarantee. \suppsec{C} records the available scaling-run configuration provenance.

\section{Discussion}
\label{sec:discussion}

BudgetMatch, SDFT, LoRA, FAPM, and CPT+SFT intervene at different points: QA repetition, data recipe, parameter-efficient adaptation, pruning-based recovery, or raw-document modeling. \method{} should therefore be read as a decomposition of adaptation budgets, not as a recipe that dominates every baseline. These controls also show why domain acquisition and general retention must be evaluated separately rather than collapsed into one score.

Recover also changes the selection object: the highest-domain IA checkpoint need not be the best deployable point because domain gains can accompany instruction-following or general-benchmark loss. We select from a small domain-general frontier using domain accuracy as the primary objective and IFEval, MMLU, and MSBench as guardrails; \suppsec{E} gives the threshold, guardrail, and tie-break rule. Boundary cases such as Phi CCI and Llama CC show why deployment may favor domain accuracy, general retention, or both.

This separation is also why we report the three general benchmarks individually. A single average can conceal whether recovery comes from instruction following, broad factual reasoning, or judged response quality. The component metrics expose those differences and make the selected operating point auditable rather than reducing recovery to one composite score.

\section{Conclusion}

We presented \method{}, separating document exposure, QA alignment, and post-hoc recovery for retrieval-free internalization. Before Recover, Inject+Align contributes domain gains across all eight main settings, although the best Inject mixture remains model- and corpus-dependent. Recover then moves these adapted checkpoints toward stronger general performance while retaining most of their domain gain, including the repeated Qwen3 scaling pattern. Against BudgetMatch, \method{} improves domain and mean general performance in three of four settings and all four reported metrics for both CCI models. These results support staged exposure and recovery as a strong, setting-dependent operating point rather than a uniformly dominant recipe.

\clearpage
\raggedbottom
\bibliography{custom}
\clearpage
\appendix
\setcounter{secnumdepth}{1}
\section*{Supplementary Material}
The following sections provide the complete supplementary methods, protocols, results, and analyses referenced by the main paper.


\section{Dataset and Task Examples}
\label{app:data-details}

CC and CCI are document-derived QA evaluations. CC is derived from Common Corpus \citep{commoncorpus}; CCI is derived from the CCI dataset \citep{cci}. In both cases, training examples are generated from source documents and test examples are held out. At inference time, all evaluated models receive only the question, not the source document or a retrieved passage. This makes the task deliberately stricter than RAG-style document QA. Table~\ref{tab:dataset-detail} summarizes the train/test contract used by both datasets.

\begin{table*}[!t]
\centering
\footnotesize
\setlength{\tabcolsep}{4pt}
\begin{tabular*}{\textwidth}{@{\extracolsep{\fill}}lllp{0.23\textwidth}p{0.14\textwidth}p{0.20\textwidth}@{}}
\toprule
Dataset & Split & Count & Input at train time & Input at test time & Primary use \\
\midrule
\multirow{2}{*}{CC} & train & 14,258 QA & question, answer, derived document fields & question only & mixed-domain internalization \\
 & test & 750 QA & \na{} & question only & held-out domain test \\
\multirow{2}{*}{CCI} & train & 10,926 QA & question, answer, derived document fields & question only & Chinese-domain internalization \\
 & test & 575 QA & \na{} & question only & held-out domain test \\
\bottomrule
\end{tabular*}
\caption{Dataset construction contract. The train/test split and retrieval-free input column define the experimental setting: models must answer from internalized parameters rather than from retrieved source passages. The files named \texttt{eval\_750.jsonl} and \texttt{eval\_575.jsonl} in the repository are treated as held-out test files in this paper.}
\label{tab:dataset-detail}
\end{table*}

\paragraph{Example prompt shape.}
The actual examples vary by corpus, but every row follows the same retrieval-free evaluation contract:
\begin{center}
\fbox{\begin{minipage}{0.92\linewidth}
\small
\textbf{Source fragment used during data construction:} A bounded corpus document contains a target fact.\\
\textbf{Generated training pair:} Question: Which fact is stated in the document? Answer: The target fact.\\
\textbf{Evaluation input:} Question only. No source passage or retrieved context is provided.\\
\textbf{Scoring:} The model answer is judged for correctness and the same checkpoint is evaluated on IFEval, MMLU, and MSBench.
\end{minipage}}
\end{center}
This box is schematic and summarizes the input contract used by all domain evaluations.

\section{Prompt Templates and QA Accounting}
\label{app:prompts}

The implementation uses Chinese instruction templates. For reproducibility, we report English prompt schemas that preserve the operative constraints, placeholders, and output contracts. The source templates are grouped into three parts: document-derived QA construction, post-training prompts, and evaluation prompts.

\paragraph{Document-derived QA construction.}
CC and CCI QA pairs are generated by an anchor-aware file-to-QA pipeline. The referenced configuration uses anchor-aware question generation and single-model answer generation. It first extracts referable anchors from each document chunk, selects applicable question types, generates self-contained questions, validates them, and then generates answers grounded in the same chunk. Table~\ref{tab:qa-generation-prompts} reports the QA-construction prompt schemas.

\begin{table*}[!t]
\centering
\footnotesize
\setlength{\tabcolsep}{4pt}
\begin{tabular}{p{0.16\linewidth}p{0.56\linewidth}p{0.20\linewidth}}
\toprule
Step & Prompt schema & Output contract \\
\midrule
Anchor extraction & Given a text chunk, extract at most $K$ independently referable core objects. Prefer explicit concepts, methods, mechanisms, modules, devices, or technical terms appearing in the text. Do not output deictic objects such as ``this method'' or ``the above mechanism.'' & JSON list of anchors. \\
Type applicability & Given a question-type description and the text chunk, decide whether the chunk can support a question of that type. The supported types are factual extraction, mechanism explanation, design rationale, condition/constraint, limitation/trade-off, and comparison/relation. & yes/no. \\
Question generation & Given the chunk, a selected anchor, a question type, the type description, and an expected answer schema, generate one natural question. The anchor must be explicitly named; the question must be understandable without the source document; deictic expressions such as ``this,'' ``above,'' or ``according to the text'' are forbidden; output only one question ending with a question mark. & Plain question text. \\
Question validation & Check whether the question is independently understandable, avoids document/deictic references, is semantically clear, and has an answer direction. & JSON with \texttt{valid} and \texttt{reason}. \\
Answer generation & Given the source chunk and generated question, answer strictly from the chunk. The answer must be faithful, accurate, professional, directly answer the question, avoid document/deictic references, and be written as a natural paragraph rather than a template with section headings. & JSON with \texttt{answer}. \\
\bottomrule
\end{tabular}
\caption{Document-derived QA construction prompts. The schemas show how chunks are converted into self-contained questions and grounded answers while preventing deictic questions that require access to the original document.}
\label{tab:qa-generation-prompts}
\end{table*}

\paragraph{Generation-stage accounting.}
The pipeline does not retain every source record or generated candidate. Table~\ref{tab:qa-generation-quality} reports micro-aggregated counts from the frozen generation artifacts. CC stage counts come from trusted run statistics; the CCI counts were recovered from validated timestamped per-domain logs and resume artifacts after the pipeline's persisted statistics files were found to reflect stale skip-rerun states. The final column reports the QA rows selected by the downstream dataset-construction step for training and held-out testing, rather than an additional quality-filter rate. Because CC and CCI use different source-sampling and prefiltering paths, their stage rates characterize data flow rather than a directly comparable dataset-quality score.

\begin{table*}[!t]
\centering
\footnotesize
\setlength{\tabcolsep}{4pt}
\begin{tabular}{lrrrrrr}
\toprule
Dataset & Input docs & Files filtered & Chunks kept & Valid questions & QA kept & Experiment QA \\
\midrule
CC  & 4,001 & 1,944 (48.6\%) & 5,324/6,593 (80.8\%) & 37,397/77,224 (48.4\%) & 16,674/37,397 (44.6\%) & 15,008 \\
CCI & 7,000 & 310 (4.4\%)   & 11,407/11,793 (96.7\%) & 100,096/140,148 (71.4\%) & 62,670/100,096 (62.6\%) & 11,501 \\
\bottomrule
\end{tabular}
\caption{QA-generation stage accounting. ``Files filtered'' gives the count and percentage of input documents rejected at file-level filtering; the chunk, question, and QA columns give retained/considered counts and micro rates. ``Experiment QA'' is the train-plus-test total after exact-question deduplication and fixed chunk-group sampling.}
\label{tab:qa-generation-quality}
\end{table*}

The archived logs expose two implementation details that are otherwise hidden by aggregate counts. Under the question validator's fail-open rule, 33 CC and 12 CCI candidates were retained after validation-response parsing failures; no candidate was retained after a validation-call error. Deterministic checks over the 26,509 final experiment rows found no malformed records, missing required fields, or normalized duplicate QA pairs. CC contains one repeated normalized question and one deictic-pattern match (``according to the text''); CCI contains neither. We also archive character-trigram answer--source overlap as a descriptive mismatch diagnostic, but do not interpret lexical overlap as correctness or faithfulness because answers may paraphrase the source or differ in script. These checks are not a substitute for human validation. Source hashes, group-level rates, and diagnostics are stored in the code package under \path{artifacts/qa_generation_quality/reports/}.

\paragraph{Inject objectives and prompts.}
Before training, source documents are converted into the three supervised document-generation datasets defined in Table~\ref{tab:inject-objectives}. The cleaning prompt asks a model to repair OCR and formatting noise while preserving terminology, numbers, formulas, symbols, style, and key details. For Rewrite, the skeleton prompt preserves entities, definitions, values, clauses, formulas, and logical chains while avoiding long direct copying; the outline prompt covers every paragraph and preserves names, places, times, data, and model names.

\begin{table*}[!t]
\centering
\footnotesize
\setlength{\tabcolsep}{3pt}
\begin{tabular}{p{0.13\linewidth}p{0.22\linewidth}p{0.19\linewidth}p{0.16\linewidth}p{0.08\linewidth}p{0.15\linewidth}}
\toprule
Objective & User input $u$ & Assistant target $y$ & Masked tokens & Recipe role & Intended exposure \\
\midrule
Continuation & Continue/complete instruction plus a document prefix. & Held-out suffix. & System prompt, instruction, and prefix. & single or mixed & Prefix-conditioned document exposure. \\
Rewrite & Reconstruction instruction plus a generated summary, outline, or knowledge skeleton. & Full cleaned document. & System prompt, instruction, and compressed representation. & single or mixed & Recover document content from a compressed representation. \\
Instruction-formatted reconstruction & Short generic reading instruction. & Full cleaned document. & System prompt and instruction. & 1:0:0 or mixed & Dense exposure through a full-document target. \\
\bottomrule
\end{tabular}
\caption{Inject objective definitions. All three objectives use the instruction model's chat template and assistant-target loss; the loss mask excludes every system/user token. Recipe ratios control the relative counts of the three objective streams, while realized shares can differ slightly after tokenizer-specific length filtering.}
\label{tab:inject-objectives}
\end{table*}

The mixture constructor samples each recipe dataset in the stated integer ratio before shuffling; Table~\ref{tab:selected-inject-settings} reports realized post-filtering counts for selected runs. Align and Vanilla instead use a question-only user prompt and apply loss only to the answer span. Context-QA diagnostics wrap source context in the user prompt but are not part of retrieval-free evaluation.

\paragraph{Evaluation prompts.}
All domain evaluations use retrieval-free model inference: the evaluated model receives the question only. Domain scoring then uses the correctness mode of the V2 evaluator. The judge receives the question, reference answer, and model answer; it is instructed to assess whether the model's core conclusion is semantically equivalent to the reference answer, ignoring source attribution. The allowed scores are 1.0 for a correct core conclusion, 0.5 for a broadly correct but incomplete or partially flawed answer, and 0.0 for an incorrect or contradictory answer. The required output is a JSON object with \texttt{score} and a short \texttt{reason}.

For general benchmarks, IFEval uses the original instruction prompt with a generic helpful-assistant system message and rule-based instruction-following scoring. MMLU uses a multiple-choice prompt and a system instruction requiring only the option letter as output. MSBench uses model inference with a helpful-assistant system prompt, then applies an LLM judge that receives the user question, reference answer, and model answer and returns JSON fields for correctness and quality.

\section{Training and BudgetMatch Details}
\label{app:training-details}

For the completed 3B/4B gradient-training runs and the extended CC baselines, serialized training arguments and original logs recover the common configuration in Table~\ref{tab:shared-optimization}. Full-parameter runs use DeepSpeed ZeRO-2 without CPU or NVMe offload; LoRA uses the same optimizer schedule while updating adapters only. All objectives mask prompt tokens and optimize the assistant target span. Table~\ref{tab:method-settings} records stage-specific settings and exceptions.

\paragraph{Computing environment.}
The archived main training runs used Linux x86\_64 nodes (glibc 2.35; kernels \texttt{5.4.0-113-generic} or \texttt{5.15.0-1053-nvidia}) with eight NVIDIA A100-SXM4-40GB GPUs, 128 physical CPU cores (256 logical cores), and approximately 1 TiB of host memory. The software stack used CPython 3.10.0, CUDA 12.4, PyTorch 2.6.0, Transformers 4.57.1, DeepSpeed 0.14.3, and PEFT 0.12.0. Model inference used vLLM 0.8.4, and Recover artifacts record mergekit 0.1.3. Logged inference jobs used one GPU unless otherwise specified; domain QA scoring called external LLM-judge APIs.

\begin{table*}[!t]
\centering
\footnotesize
\setlength{\tabcolsep}{5pt}
\begin{tabular}{p{0.16\textwidth}p{0.29\textwidth}p{0.16\textwidth}p{0.29\textwidth}}
\toprule
Parameter & Value & Parameter & Value \\
\midrule
Optimizer & AdamW (PyTorch) & Learning rate & $5\times10^{-5}$ \\
Adam $\beta_1,\beta_2,\epsilon$ & $.9,.999,10^{-8}$ & Scheduler / warmup & cosine / ratio .05 \\
Weight decay / max grad norm & $.01 / 1.0$ & Precision / max length & BF16 / 4096 \\
Batch per GPU & 1 & Gradient accum. / GPUs & $8 / 8$ \\
Effective global batch & 64 examples/step & DeepSpeed & ZeRO-2, no offload \\
Termination / checkpoint & epoch based / final epoch & Gradient checkpointing & model-level enabled \\
\bottomrule
\end{tabular}
\caption{Shared optimization settings recovered for the completed 3B/4B training runs and extended CC baselines. The effective global batch is per-device batch $1\times8$ accumulation steps $\times8$ GPUs.}
\label{tab:shared-optimization}
\end{table*}

\begin{table*}[!t]
\centering
\footnotesize
\setlength{\tabcolsep}{4pt}
\begin{tabular}{p{0.16\textwidth}rp{0.31\textwidth}p{0.38\textwidth}}
\toprule
Method / stage & Epochs & Data and objective & Method-specific setting \\
\midrule
Vanilla SFT & 3 & QA; answer-only & Original Instruct initialization \\
BudgetMatch & 14/17/21/11 & Same QA and loss as Vanilla & CC Llama/Qwen; CCI Llama/Qwen order \\
Inject & 3 & Three assistant-target document-generation objectives & Selected mixtures and counts in Table~\ref{tab:selected-inject-settings} \\
Align & 3 & QA; answer-only & Initializes from the Inject final epoch \\
SDFT & 3 & Model-specific synthetic QA; answer-only & 14,258 CC examples per model \\
LoRA & 3 & CC QA; answer-only & $r=16$, $\alpha=32$, dropout .05; merged for evaluation \\
Replay & 3 & 75\% domain QA + 25\% general instruction & Equal-size replacement; construction seed 42 \\
CPT & 16 & Raw-document causal LM & Matched Base initialization; 4,425 CC / 10,769 CCI rows \\
CPT+SFT & 3 & QA; answer-only & Initializes from the CPT final epoch \\
\bottomrule
\end{tabular}
\caption{Stage-specific training settings. Unless stated as an exception, each row uses Table~\ref{tab:shared-optimization}. BudgetMatch epoch counts are setting-specific rather than a shared 13-epoch approximation.}
\label{tab:method-settings}
\end{table*}

\begin{table*}[!t]
\centering
\footnotesize
\setlength{\tabcolsep}{5pt}
\begin{tabular}{llrrr}
\toprule
Dataset & Model & Selected Inject recipe & Inject rows & Align QA rows \\
\midrule
\multirow{4}{*}{CC} & Llama-3.2-3B & Mixed 1:1:2 & 19,000 & 14,258 \\
 & Phi-4-mini & Mixed 1:1:2 & 19,000 & 14,258 \\
 & Qwen3-4B & Mixed 1:1:1 & 18,968 & 14,258 \\
 & SmolLM3-3B & Mixed 1:1:1 & 18,960 & 14,258 \\
\midrule
\multirow{4}{*}{CCI} & Llama-3.2-3B & Mixed 1:1:2 & 19,000 & 10,926 \\
 & Phi-4-mini & Mixed 1:1:2 & 19,000 & 10,926 \\
 & Qwen3-4B & Reconstruction 1:0:0 & 10,000 & 10,926 \\
 & SmolLM3-3B & Mixed 1:1:2 & 19,000 & 10,926 \\
\bottomrule
\end{tabular}
\caption{Selected Inject configurations for the eight main settings. Counts are realized post-tokenization training rows; tokenizer-specific length filtering explains the small differences among nominally equal mixtures.}
\label{tab:selected-inject-settings}
\end{table*}

\paragraph{BudgetMatch token accounting.}
The completed-run accounting accumulates non-padding training tokens under each model's tokenizer. We define IA total as Inject plus Align and compare it directly with the realized BudgetMatch token volume. The integer training schedules---14, 17, 21, and 11 epochs for CC Llama, CC Qwen3-4B, CCI Llama, and CCI Qwen3-4B---are implementation settings reported in Table~\ref{tab:method-settings}. Matching is assessed from the realized token volumes, so no separate fractional-epoch estimate is reported. Table~\ref{tab:budget-accounting} reports these volumes rounded to 0.001 million.

\begin{table*}[!t]
\centering
\footnotesize
\setlength{\tabcolsep}{4pt}
\begin{tabular*}{\textwidth}{@{\extracolsep{\fill}}llrrrrr@{}}
\toprule
Setting & Inject recipe (samples) & Inject & Align & IA total & BudgetMatch & BM/IA \\
\midrule
CC Llama-3.2-3B & Mixed 1:1:2 (19k) & 45.736 & 12.917 & 58.653 & 60.280 & 102.8\% \\
CC Qwen3-4B & Mixed 1:1:1 (18,968) & 46.017 & 10.191 & 56.208 & 57.752 & 102.7\% \\
CCI Llama-3.2-3B & Mixed 1:1:2 (19k long-doc.) & 57.296 & 9.734 & 67.030 & 68.085 & 101.6\% \\
CCI Qwen3-4B & 1:0:0 (10k) & 19.081 & 7.151 & 26.232 & 26.202 & 99.9\% \\
\bottomrule
\end{tabular*}
\caption{Realized token accounting for BudgetMatch, in millions of non-padding model tokens. BM/IA compares the completed QA-only BudgetMatch run directly with the corresponding Inject+Align token volume.}
\label{tab:budget-accounting}
\end{table*}

\paragraph{Runs and training seeds.}
Every reported checkpoint is a single training run; we do not average over multiple seeds. Hugging Face \texttt{TrainingArguments} uses \texttt{seed=42} and \texttt{data\_seed=42}. The tokenized training data are shuffled with seed 1234, and the internal training/validation split uses seed 42. Dataloader \texttt{drop\_last} is enabled. DeepSpeed runs save once per epoch and do not load a validation-best checkpoint; the reported \texttt{final\_model} is the final-epoch model. Gradient checkpointing is enabled by a direct model-level call even though the serialized \texttt{TrainingArguments} flag is false. These fixed seeds improve run traceability but do not guarantee bitwise determinism under distributed GPU training.

\paragraph{Scaling-run parameter provenance.}
The selected Qwen3-8B/14B/32B Recover configurations are retained and all use TIES density .3, but the original Inject/Align training arguments, logs, and node manifests for the scaling ablation are not present in the available run archive. We therefore do not infer their LR, batch, sequence length, precision, GPU count/model, node count, or ZeRO/offload settings from current launcher defaults. The computing-environment paragraph and Table~\ref{tab:shared-optimization} should not be read as covering these three scaling rows.

\section{Evaluation and Judge Reliability}
\label{app:evaluation-details}

Domain accuracy is produced by the repository's V2 evaluator. The evaluator records model generations, per-example judge decisions, and aggregate correctness. The general benchmark suite contains IFEval, MMLU, and MSBench. IFEval and MMLU are rule- or answer-key-based in the current pipeline; MSBench uses an LLM judge and therefore shares the audit requirements of domain QA. Table~\ref{tab:metric-provenance} defines the metric directions and evaluation types.

MSBench is public: we use a fixed 200-example local evaluation subset drawn from the Apache-2.0 \texttt{iic/ms\_bench} release, itself a public subset of MSAgent-Bench \citep{li2023modelscope,msbench}. The source release provides the conversational examples; the fixed 200-example selection, model prompting, and adaptive LLM-judge aggregation described below are our repository-level evaluation protocol rather than a claim of reproducing a separate canonical leaderboard.

\begin{table*}[!t]
\centering
\footnotesize
\begin{tabular}{llll}
\toprule
Metric & Direction & Evaluation type & Reported unit \\
\midrule
Domain accuracy & higher is better & V2 multi-judge correctness & percentage \\
IFEval inst-strict & higher is better & instruction-following evaluator & percentage \\
MMLU accuracy & higher is better & multiple-choice benchmark & percentage \\
MSBench accuracy & higher is better & LLM-judge benchmark & percentage \\
\bottomrule
\end{tabular}
\caption{Evaluation metrics and their roles in operating-point selection. Domain QA is the primary retrieval-free internalization metric, while IFEval, MMLU, and MSBench serve as general-capability guardrails.}
\label{tab:metric-provenance}
\end{table*}

\paragraph{Inference and judge decoding.}
Each checkpoint is evaluated with one model-generation pass and one judging pass. Domain QA shuffles or truncates the test input with seed 42, then generates with temperature .7, \texttt{top\_p=.95}, repetition penalty 1.1, and a 2,048-token limit. No sampler seed is passed to vLLM, so the domain generation remains stochastic. The reported non-thinking IFEval, MMLU, and MSBench runs use greedy decoding with token limits of 1,024, 10, and 1,024, respectively; their optional subsampling order uses seed 123. MSBench judge calls override temperature to .1.

Judge APIs receive no explicit random seed. Their stored configurations use temperature/top-$p$ pairs of .7/.8 for \texttt{gpt-oss-120b}, .6/.4 for local \texttt{MiniMax2.5}, and .7/.95 for cloud DeepSeek variants; the configured output limits are 131,072, 196,608, and 16,384 tokens, respectively. Consequently, fixed training and data-order seeds should not be read as repeated-run control over stochastic domain generation or API judging.

\paragraph{Judge prompt contract.}
The domain and MSBench judges share the same evidence inputs, but use different output and aggregation rules:
\begin{center}
\fbox{\begin{minipage}{0.92\linewidth}
\small
\textbf{Input:} question, reference answer, model answer, optional scoring rubric.\\
\textbf{Decision:} domain QA returns a score in $\{0,.5,1\}$; MSBench returns binary correctness and a quality score.\\
\textbf{Required rationale:} short explanation identifying whether the model answer contains the required fact and whether it introduces unsupported content.\\
\textbf{Domain aggregation:} two judges score every answer; if their scores differ, a third judge is called. The final score is the median of the available scores, and every raw vote is stored before aggregation.
\end{minipage}}
\end{center}
MSBench also calls two judges first and a third on binary disagreement; final correctness uses majority vote and quality uses the mean of available quality scores. These contracts are part of the evaluation method because both metrics are judge-produced rather than answer-key labels.

\paragraph{Judge agreement audit.}
The audit reads stored per-example V2 JSONL files without calling a model or judge. It covers 343 result files and 242,255 valid model-answer records; ten malformed JSON lines are skipped, and every valid record contains raw correctness votes. Because the panel is adaptive, agreement is measured on the first two judges and the third-judge rate measures arbitration frequency.

\begin{table*}[!t]
\centering
\footnotesize
\setlength{\tabcolsep}{5pt}
\begin{tabular}{lrrrrr}
\toprule
Scope & Records & Exact agree & Binary agree & Binary $\kappa$ & Third judge \\
\midrule
CC test artifacts & 163,347 & .732 & .854 & .691 & .274 \\
CCI test artifacts & 78,308 & .656 & .836 & .670 & .345 \\
Full audit & 242,255 & .707 & .848 & .691 & .297 \\
\bottomrule
\end{tabular}
\caption{Domain QA judge reliability. Binary agreement collapses scores at $\geq .5$. The full audit additionally contains 600 samples from a smaller CCI evaluation set. These statistics quantify judge reliability over the evaluated samples.}
\label{tab:judge-audit}
\end{table*}

\begin{table*}[!t]
\centering
\footnotesize
\setlength{\tabcolsep}{7pt}
\begin{tabular}{lrrrr}
\toprule
Judge & Votes & Mean score & $P(s\geq .5)$ & $P(s=1)$ \\
\midrule
\texttt{gpt-oss-120b} & 242,031 & .387 & .454 & .321 \\
\texttt{MiniMax2.5\_Local} & 156,812 & .303 & .434 & .172 \\
\texttt{deepseek-v3.2} & 131,172 & .296 & .465 & .127 \\
\texttt{deepseek-v3.1} & 24,697 & .378 & .652 & .103 \\
\bottomrule
\end{tabular}
\caption{Per-judge score distributions before aggregation. Vote counts differ because panel configurations vary across evaluations and later judges are invoked adaptively; the marginal means are therefore descriptive and should not be interpreted as a controlled judge ranking.}
\label{tab:per-judge}
\end{table*}

\paragraph{Bootstrap uncertainty and audit artifacts.}
For every result file, the audit performs 2,000 example-level bootstrap resamples with seed 20260706 and reports percentile 95\% intervals for domain accuracy and mean correctness. The full per-file intervals, per-judge scores, pairwise agreement, and dataset summaries are stored in the code package under \path{artifacts/domain_eval_reliability/}. These intervals quantify evaluation-sample uncertainty for each checkpoint and support the reported domain estimates; training-seed robustness remains outside their scope.

\section{Recover Selection Protocol}
\label{app:selection}

The \method{} rows use a domain-primary frontier criterion on the training-framework validation split. Recover candidates are first filtered for meaningful domain accuracy relative to Vanilla SFT, then compared on IFEval, MMLU, and MSBench as guardrails. The selected row is therefore an operating point, not necessarily the maximum-domain checkpoint. The merge candidate is fixed before running the held-out test files used in the main tables. Recover is implemented as a selection mechanism over existing post-hoc weight-space merge operators, not as a new merging algorithm.

The repository uses \emph{eval} in two different places. During training and Recover selection, eval denotes an internal validation split derived from the training data. The published CC and CCI files with 750 and 575 examples are held-out test sets, despite their local \texttt{eval\_*.jsonl} filenames. We use \emph{test} for those files throughout the paper to avoid implying that final reported scores were also used for Recover-candidate selection.

\paragraph{Formal selection rule.}
For each dataset--model setting, let $c$ denote a Recover candidate and let $v$ denote the corresponding Vanilla SFT checkpoint on the validation split. We write $D(c)$ for retrieval-free domain QA accuracy, $I(c)$ for IFEval, $M(c)$ for MMLU, $B(c)$ for MSBench, and
\[
G(c)=\frac{I(c)+M(c)+B(c)}{3}
\]
for mean general performance. We use a fixed tolerance of $\tau=1.0$ percentage point. Candidate selection proceeds as follows:
\begin{enumerate}
\item \textbf{Domain feasibility.} Keep candidates with $D(c)\geq D(v)-\tau$. This allows boundary trade-offs such as Phi CCI only when the domain loss relative to Vanilla SFT is within tolerance.
\item \textbf{General guardrail.} Among domain-feasible candidates, keep candidates with $G(c)\geq G(v)$ and at least two of $\{I,M,B\}$ no more than $\tau$ below the corresponding Vanilla SFT value.
\item \textbf{Domain-primary ranking.} Select from the non-dominated candidates in the $(D,G)$ plane. Domain accuracy is the primary key; candidates within $\tau$ of the best remaining domain score are treated as the same domain tier.
\item \textbf{Tie-break.} Within the best domain tier, choose the candidate with the largest $G(c)$. Remaining ties are broken by the larger minimum improvement over Vanilla SFT across $\{I,M,B\}$, then by the smaller merge hyperparameter within the same operator family.
\end{enumerate}
This rule makes Recover a validation-time operating-point selector rather than an unreported search over the held-out test results.

\begin{figure*}[!t]
\centering
  \includegraphics[width=0.98\textwidth]{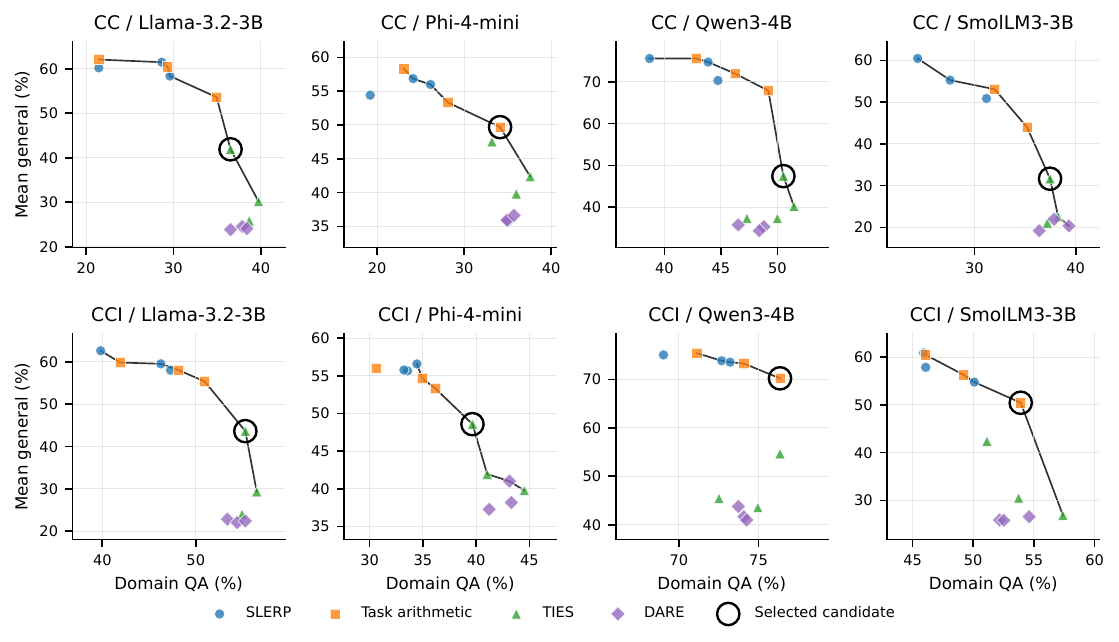}
  \caption{Recover candidate frontiers for the main dataset--model settings. Each panel plots held-out test performance of the fixed Recover candidates by retrieval-free domain QA accuracy and mean general performance across IFEval, MMLU, and MSBench. The black line marks non-dominated points in this two-dimensional projection, and the black ring highlights the candidate selected by the validation protocol for the main \method{} table. The displayed test frontier is diagnostic and is not used to choose the selected candidate.}
  \label{fig:selection-frontiers}
\end{figure*}

Table~\ref{tab:recover-grid} gives the fixed candidate grid used for each recoverable IA checkpoint. Figure~\ref{fig:selection-frontiers} visualizes the candidate frontier for each dataset--model setting, and Table~\ref{tab:selected-recovery-settings} lists the selected operating point for each main \method{} row.

\begin{table*}[!t]
\centering
\footnotesize
\begin{tabular}{lll}
\toprule
Operator family & Hyperparameter grid & Candidate count \\
\midrule
SLERP & $t\in\{0.2,0.3,0.4\}$ & 3 \\
Task Arithmetic & $w\in\{0.3,0.5,0.7\}$ & 3 \\
TIES & $d\in\{0.3,0.5,0.7\}$ & 3 \\
DARE & $d_r\in\{0.1,0.3,0.5\}$ & 3 \\
\midrule
Total & fixed grid per IA checkpoint & 12 \\
\bottomrule
\end{tabular}
\caption{Recover candidate grid. Every selected \method{} row is chosen from this fixed set of post-hoc merge candidates rather than from an unreported per-row search space. Candidate choice is made on the validation split before held-out test reporting.}
\label{tab:recover-grid}
\end{table*}

All Recover outputs use BF16 weights and the base-model tokenizer. TIES fixes task-vector weight to 1.0. DARE-TIES converts drop rate $d_r$ to density $1-d_r$, fixes task-vector weight to 1.0, and enables normalization. For the selected CC Phi Recover checkpoint, Task Arithmetic applies $w=0.7$ to the full task vector; the corresponding source and output-local configurations are retained with matching hashes in the code supplement. For tied-embedding models, the pipeline temporarily materializes the LM head for merging and restores the original tying afterward.

\begin{table*}[!t]
\centering
\footnotesize
\setlength{\tabcolsep}{4pt}
\begin{tabular*}{\textwidth}{@{\extracolsep{\fill}}lllrrrrp{0.18\textwidth}@{}}
\toprule
Dataset & Model & Selected checkpoint & Domain (\%) & IFEval (\%) & MMLU (\%) & MSBench (\%) & Selection note \\
\midrule
\multirow{4}{*}{CC} & Llama-3.2-3B & TIES $d=0.3$ & 36.5 & 60.2 & 35.0 & 30.5 & domain-primary feasible point \\
 & Phi-4-mini & Task Arithmetic $w=0.7$ & 34.1 & 49.0 & 57.0 & 43.0 & balanced feasible point \\
 & Qwen3-4B & TIES $d=0.3$ & 50.5 & 59.8 & 19.5 & 63.0 & domain-primary feasible point \\
 & SmolLM3-3B & TIES $d=0.3$ & 37.5 & 40.3 & 25.7 & 29.0 & domain-primary feasible point \\
\midrule
\multirow{4}{*}{CCI} & Llama-3.2-3B & TIES $d=0.3$ & 55.3 & 61.3 & 33.2 & 36.5 & domain-primary feasible point \\
 & Phi-4-mini & TIES $d=0.3$ & 39.7 & 51.6 & 50.2 & 44.0 & boundary trade-off point \\
 & Qwen3-4B & Task Arithmetic $w=0.7$ & 76.3 & 76.1 & 64.5 & 70.0 & domain-primary feasible point \\
 & SmolLM3-3B & Task Arithmetic $w=0.7$ & 53.9 & 57.4 & 46.8 & 47.0 & domain-primary feasible point \\
\bottomrule
\end{tabular*}
\caption{Selected Recover settings for the main \method{} rows. The table reports held-out test scores for the concrete operating point chosen from the fixed Recover grid by the validation protocol, making the domain-primary selection rule explicit rather than treating recovery as an unreported hyperparameter search.}
\label{tab:selected-recovery-settings}
\end{table*}

\section{Qwen3-4B CCI Diagnostic}
\label{app:qwen-cci-prior}

Qwen3-4B on CCI is an important boundary case because the original instruction checkpoint already reaches 70.6\% domain accuracy before any document-internalization training. This is much higher than the corresponding initial scores for Llama, Phi, and SmolLM on CCI, and it changes how the CCI Qwen3-4B rows should be interpreted. Here, the initial checkpoint is the instruction model used before any training in this work, not a pure pretrained Base checkpoint. The main question in this setting is not whether post-training can create a large absolute domain gain from a weak starting point, but whether it can preserve or improve an already strong domain prior while recovering general capability.

We therefore avoid treating the CCI Qwen3-4B recipe result as evidence that a single \inject{} mixture is universally best. Vanilla SFT reaches 75.1\%, the best pre-recovery Inject+Align row reaches 75.5\%, and the selected \method{} checkpoint reaches 76.3\% while substantially improving IFEval, MMLU, and MSBench over Vanilla SFT. The small pre-recovery domain gap is consistent with a high-base-prior setting: there is limited headroom for document exposure to improve domain accuracy, so the value of \recover{} is more visible in the domain-general trade-off than in raw domain gain.

We conduct a post-hoc diagnostic on the four original instruction checkpoints to determine whether Qwen3-4B assigns higher likelihood to the evaluated CCI source-text distribution. The CCI test file contains 575 QA rows, which map to 509 unique source documents after normalization for document identity; scoring always preserves the original raw text. We score source text only, without questions, answers, special tokens, or chat templates. For each document, we select a UTF-8 prefix--target boundary near the byte midpoint, prefer a sentence or paragraph boundary within the 45--55\% range, require at least 128 bytes on each side, and require that the boundary be valid under the canonical full-text tokenization of all four models. This produces 507 shared continuations; two short documents are excluded.

Because token-level perplexity is not comparable across different tokenizers, we report conditional bits per byte:
\[
\mathrm{BPB}=\frac{\sum_t -\log p(x_t\mid x_{<t})}{\ln 2\cdot B},
\]
where $B$ is the number of UTF-8 bytes in the shared target. All models predict exactly the same target bytes with deterministic teacher-forced scoring in BF16, using a maximum context length of 4096 and a stride of 2048; overlapping windows provide context only, and each target token is scored once. For all four models, a 20-document smoke test agreed with direct single-window forward computation to within $6.1\times 10^{-7}$, below the $10^{-5}$ tolerance. For each peer, we compute paired document-level Qwen-minus-peer BPB differences. The primary statistic averages the mean difference within each of the seven CCI domains and then weights domains equally; 95\% intervals use 10,000 paired bootstrap samples stratified by domain. Document-weighted and byte-weighted aggregates are secondary summaries.

Qwen3-4B has lower equal-domain macro BPB than every peer, with all paired 95\% bootstrap intervals below zero and all seven domain means in the same direction (Table~\ref{tab:qwen-cci-bpb}).

\begin{table*}[!t]
\centering
\footnotesize
\setlength{\tabcolsep}{5pt}
\begin{tabular*}{\textwidth}{@{\extracolsep{\fill}}lrrrp{0.27\textwidth}@{}}
\toprule
Initial checkpoint & CCI QA (\%) & Doc. mean BPB & Corpus BPB & Qwen--peer macro $\Delta$ [95\% CI] \\
\midrule
Qwen3-4B-Instruct & 70.6 & 0.744 & 0.729 & -- \\
Llama-3.2-3B-Instruct & 29.2 & 1.021 & 0.998 & $-0.273$ [$-0.285$, $-0.261$] \\
Phi-4-mini-instruct & 27.3 & 0.968 & 0.948 & $-0.220$ [$-0.231$, $-0.208$] \\
SmolLM3-3B & 34.1 & 0.838 & 0.819 & $-0.093$ [$-0.102$, $-0.084$] \\
\bottomrule
\end{tabular*}
\caption{Conditional BPB on identical CCI source-document continuations. Corpus BPB is byte weighted; the primary effect first averages paired Qwen-minus-peer BPB differences within each of seven domains and then weights domains equally. Negative $\Delta$ means lower BPB for Qwen. All intervals exclude zero and all seven domain means agree in direction.}
\label{tab:qwen-cci-bpb}
\end{table*}

The difference is broad at the document level: Qwen has lower BPB than Llama on 100\% of documents, Phi on 99.0\%, and SmolLM3 on 86.8\%. This pattern is consistent with stronger fit to the evaluated CCI text distribution as one possible contributor to Qwen's high initial QA accuracy. It is not a sufficient explanation of QA behavior: for example, Phi has lower BPB than Llama but slightly lower CCI QA accuracy. Moreover, the diagnostic compares complete checkpoints rather than controlling model capacity, pretraining composition, or Chinese-language capability. It therefore does not identify why the likelihood difference arose, establish that text likelihood causes the QA gap, or demonstrate memorization, training-data overlap, or contamination.

The public Qwen3 technical report does not report this CCI QA benchmark, but it does disclose strong base-model performance for the Qwen3 dense family across general and multilingual evaluations \citep{qwen3technicalreport}. Table~\ref{tab:qwen3-official-family} records the relevant family-level numbers. These official results do not establish CCI overlap, but they make a high-base-prior explanation plausible: Qwen3 starts from a strong multilingual and knowledge benchmark profile before any document-internalization training in our experiments.

\begin{table*}[!t]
\centering
\footnotesize
\begin{tabular}{lrrrrrr}
\toprule
Model & MMLU & MMLU-Pro & BBH & MGSM & MMMLU & INCLUDE \\
\midrule
Qwen3-1.7B-Base & 62.63 & 36.76 & 54.47 & 50.71 & 63.27 & 45.57 \\
Qwen3-4B-Base & 72.99 & 50.58 & 72.59 & 67.74 & 71.42 & 56.29 \\
Qwen3-8B-Base & 76.89 & 56.73 & 78.40 & 76.02 & 75.72 & 59.40 \\
Qwen3-14B-Base & 81.05 & 61.03 & 81.07 & 79.20 & 79.69 & 64.55 \\
Qwen3-32B-Base & 83.61 & 65.54 & 87.38 & 83.06 & 83.83 & 67.87 \\
\bottomrule
\end{tabular}
\caption{Officially disclosed Qwen3 dense-family base-model scores from the Qwen3 technical report. These are public general and multilingual benchmark results, not CCI QA results, and are included only to contextualize the high-base-prior interpretation.}
\label{tab:qwen3-official-family}
\end{table*}

Table~\ref{tab:qwen-cci-diagnostics} reports the available Qwen3 CCI diagnostic rows. The 1.7B and 4B rows both start from unusually high CCI base scores relative to non-Qwen models. For Qwen3-1.7B, post-training does not improve over the base score, suggesting that the CCI signal can already be strong in the base model and that additional domain fitting can disturb that prior. For Qwen3-4B, Vanilla SFT and the 1:0:0 Inject recipe improve the domain score modestly, but the small gap between Vanilla SFT and the best pre-recovery recipe explains why this setting should be read as a high-base-prior boundary case rather than as evidence for a universally best Inject mixture.

\begin{table*}[!t]
\centering
\footnotesize
\begin{tabular}{llr}
\toprule
Model & Setting & CCI domain (\%) \\
\midrule
\multirow{3}{*}{Qwen3-1.7B} & Base Instruct & \textbf{60.0} \\
 & Vanilla SFT & 57.0 \\
 & Mixed 1:1:1 + Stage2 & 56.7 \\
\midrule
\multirow{9}{*}{Qwen3-4B} & Base Instruct & 70.6 \\
 & Vanilla SFT & 75.1 \\
 & Context-aware SFT & 65.0 \\
 & Mixed 1:1:1 + Stage2 & 73.7 \\
 & Continue + Stage2 & 72.9 \\
 & 1:0:0 + Stage2 & \textbf{75.5} \\
 & Rewrite + Stage2 & 74.4 \\
 & Mixed 1:1:2 + Stage2 & 72.5 \\
 & Replay & 66.4 \\
\bottomrule
\end{tabular}
\caption{Qwen3 CCI high-base-prior diagnostics. The 1.7B and 4B rows show that Qwen3 starts unusually high on CCI, so this setting is better interpreted as preserving and recovering a strong prior than as creating a large new domain gain.}
\label{tab:qwen-cci-diagnostics}
\end{table*}

The same pattern appears in the Qwen3-4B Recover sweep. Table~\ref{tab:qwen-cci-recover-sweep} shows that several merge operators recover to the mid-70s domain range, with Task Arithmetic $w=0.7$ and TIES $d=0.3$ reaching 76.3\%. This is consistent with the main-table interpretation: the Qwen3-4B CCI row is not primarily a large domain-gain story, because the model starts high; its value is that Recover can preserve the high CCI prior while restoring substantially stronger general capability.

\begin{table*}[!t]
\centering
\footnotesize
\begin{tabular}{llr@{\qquad}llr}
\toprule
Operator & Coefficient & CCI domain (\%) & Operator & Coefficient & CCI domain (\%) \\
\midrule
DARE & $d_r=0.1$ & 74.1 & SLERP & $t=0.2$ & 69.0 \\
DARE & $d_r=0.3$ & 74.3 & SLERP & $t=0.3$ & 72.7 \\
DARE & $d_r=0.5$ & 73.7 & SLERP & $t=0.4$ & 73.2 \\
\midrule
Task Arithmetic & $w=0.3$ & 71.1 & TIES & $d=0.3$ & \textbf{76.3} \\
Task Arithmetic & $w=0.5$ & 74.1 & TIES & $d=0.5$ & 72.5 \\
Task Arithmetic & $w=0.7$ & \textbf{76.3} & TIES & $d=0.7$ & 75.0 \\
\bottomrule
\end{tabular}
\caption{Qwen3-4B CCI Recover sweep. Multiple merge operators stay in the high-domain range, with Task Arithmetic and TIES reaching the main selected value; this supports the boundary-case reading of the Qwen3 CCI row.}
\label{tab:qwen-cci-recover-sweep}
\end{table*}

\section{Extended Baselines and Recipe Analyses}
\label{app:baselines}
\label{app:fapm}
\label{app:recipe-grid}

\paragraph{Complete CC comparison.}
Tables~\ref{tab:cc-baselines-full} and~\ref{tab:cc-baselines-full-qwen-smol} complement the main-paper comparison with every extended-baseline row.

\begin{table*}[!t]
\centering
\footnotesize
\setlength{\tabcolsep}{5pt}
\begin{tabular*}{\textwidth}{@{\extracolsep{\fill}}llrrrr@{}}
\toprule
Model & Method & Domain (\%) & IFEval (\%) & MMLU (\%) & MSBench (\%) \\
\midrule
\multirow{7}{*}{Llama-3.2-3B} & Vanilla SFT & 35.5 & 54.2 & 11.2 & 21.5 \\
 & SDFT & 39.9 & 58.4 & 23.5 & 33.5 \\
 & LoRA & 22.5 & 64.0 & 53.2 & 30.0 \\
 & Replay & 30.8 & 56.7 & 30.7 & 37.5 \\
 & Vanilla-FAPM & 22.3 & 75.3 & 51.8 & 62.0 \\
 & IA-FAPM & 22.4 & 74.5 & 53.7 & 59.0 \\
 & \method{} & 36.5 & 60.2 & 35.0 & 30.5 \\
\midrule
\multirow{7}{*}{Phi-4-mini} & Vanilla SFT & 24.4 & 47.8 & 51.0 & 32.5 \\
 & SDFT & 32.9 & 43.2 & 58.7 & 45.5 \\
 & LoRA & 18.0 & 70.6 & 59.3 & 39.5 \\
 & Replay & 24.7 & 48.7 & 28.8 & 48.5 \\
 & Vanilla-FAPM & 17.1 & 56.4 & 58.2 & 64.5 \\
 & IA-FAPM & 16.9 & 53.8 & 61.3 & 63.0 \\
 & \method{} & 34.1 & 49.0 & 57.0 & 43.0 \\
\bottomrule
\end{tabular*}
\caption{Complete CC extended-baseline comparison for Llama-3.2-3B and Phi-4-mini.}
\label{tab:cc-baselines-full}
\end{table*}

\begin{table*}[!t]
\centering
\footnotesize
\setlength{\tabcolsep}{5pt}
\begin{tabular*}{\textwidth}{@{\extracolsep{\fill}}llrrrr@{}}
\toprule
Model & Method & Domain (\%) & IFEval (\%) & MMLU (\%) & MSBench (\%) \\
\midrule
\multirow{7}{*}{Qwen3-4B} & Vanilla SFT & 42.4 & 51.1 & 8.8 & 51.0 \\
 & SDFT & 44.1 & 55.9 & 14.0 & 53.0 \\
 & LoRA & 31.3 & 68.2 & 48.5 & 53.0 \\
 & Replay & 42.8 & 69.7 & 34.7 & 66.5 \\
 & Vanilla-FAPM & 41.2 & 83.8 & 61.3 & 83.5 \\
 & IA-FAPM & 44.1 & 82.6 & 65.5 & 85.0 \\
 & \method{} & 50.5 & 59.8 & 19.5 & 63.0 \\
\midrule
\multirow{7}{*}{SmolLM3-3B} & Vanilla SFT & 32.1 & 35.6 & 10.5 & 25.0 \\
 & SDFT & 36.5 & 49.8 & 44.3 & 25.5 \\
 & LoRA & 15.5 & 66.3 & 22.5 & 42.0 \\
 & Replay & 33.6 & 53.1 & 28.0 & 34.0 \\
 & Vanilla-FAPM & 24.7 & 71.0 & 35.5 & 68.5 \\
 & IA-FAPM & 26.1 & 69.4 & 49.0 & 67.0 \\
 & \method{} & 37.5 & 40.3 & 25.7 & 29.0 \\
\bottomrule
\end{tabular*}
\caption{Complete CC extended-baseline comparison for Qwen3-4B and SmolLM3-3B.}
\label{tab:cc-baselines-full-qwen-smol}
\end{table*}

\paragraph{Initialization diagnostic for CPT+SFT.}
The main table treats Base-initialized CPT+SFT as the conventional dense-document baseline, not as an initialization-controlled estimate of the Inject objective. Table~\ref{tab:instruct-cpt-diagnostics} therefore exposes the three completed Instruct-initialized CPT+SFT runs available in the archive. Their coverage is incomplete and they are diagnostic rather than a second main baseline matrix. On domain QA, Instruct-CPT+SFT is competitive with the best pre-Recovery IA checkpoint for CC Llama (38.7 versus 38.3), but is lower for CC Phi (31.2 versus 32.1) and CCI Llama (52.9 versus 58.6). These results show that CPT is initialization-sensitive and do not support a uniform ordering between structured Inject and raw-document CPT.

\begin{table*}[!t]
\centering
\footnotesize
\setlength{\tabcolsep}{5pt}
\begin{tabular}{llrrrr}
\toprule
Dataset & Model & Domain (\%) & IFEval (\%) & MMLU (\%) & MSBench (\%) \\
\midrule
CC  & Llama-3.2-3B & 38.7 & 43.5 & 9.8 & 19.5 \\
CC  & Phi-4-mini   & 31.2 & 43.3 & 40.7 & 31.0 \\
CCI & Llama-3.2-3B & 52.9 & 29.0 & 8.3 & 23.0 \\
\bottomrule
\end{tabular}
\caption{Available Instruct-initialized CPT+SFT diagnostics. CC Llama exposes a domain--general trade-off rather than uniform \method{} dominance, while the CC Phi and CCI Llama diagnostics remain below the corresponding best IA domain scores. Coverage is limited to completed archived runs and is not extrapolated to Qwen3 or SmolLM3.}
\label{tab:instruct-cpt-diagnostics}
\end{table*}

\paragraph{Interpretation.}
SDFT is a supervised data-recipe baseline. LoRA is a parameter-efficiency baseline that can preserve broad capability while under-internalizing the target documents. Base-initialized CPT+SFT is a conventional dense-document baseline, whereas Table~\ref{tab:instruct-cpt-diagnostics} separately shows the available initialization-controlled diagnostics. Phi-4-mini is unavailable in the Base-initialized CPT+SFT matrix because the release used here does not provide a corresponding non-instruction checkpoint.

\paragraph{Baseline implementation settings.}
SDFT trains each model from its original Instruct checkpoint on a model-specific 14,258-row synthetic CC QA file using the shared three-epoch configuration in Table~\ref{tab:shared-optimization}. LoRA uses rank 16, scaling factor 32, dropout .05, and no bias; it targets the attention and MLP projections (fused projections for Phi), and evaluation uses the merged BF16 model. Replay replaces, rather than appends, 25\% of the domain QA rows with general instructions using seed 42, preserving the original dataset size before tokenizer-specific length filtering. CPT trains for 16 epochs on raw-document causal LM and then applies the same three-epoch answer-only SFT stage. The historical launcher set this 16-epoch Stage~1 schedule to approximately match the Inject-stage token budget. Main-table CPT runs start from the corresponding Base checkpoint for Llama, Qwen, and SmolLM; the diagnostic runs in Table~\ref{tab:instruct-cpt-diagnostics} instead start from the Instruct checkpoint.

\paragraph{FAPM baseline.}
FAPM is applied as a post-hoc pruning-based recovery method at sparsity 0.9, retaining the top 10\% of task-vector entries under an independently computed per-tensor score threshold; retained entries keep the fine-tuning delta and pruned entries revert to the Instruct weight. Outputs are stored in BF16. \textit{Vanilla-FAPM} means applying FAPM to the Vanilla SFT checkpoint, while \textit{IA-FAPM} means applying the same procedure to the best pre-recovery Inject+Align checkpoint. This distinguishes the recovery mechanism from the adaptation checkpoint it starts from. Table~\ref{tab:fapm} isolates the domain effect, while Table~\ref{tab:fapm-full} reports the domain-general view.

\begin{table*}[!t]
\centering
\footnotesize
\begin{tabular}{llrrr}
\toprule
Dataset & Model & Vanilla SFT (\%) & Vanilla-FAPM (\%) & IA-FAPM (\%) \\
\midrule
\multirow{4}{*}{CC} & Llama-3.2-3B & \textbf{35.5} & 22.3 & 22.4 \\
 & Phi-4-mini & \textbf{24.4} & 17.1 & 16.9 \\
 & Qwen3-4B & 42.4 & 41.2 & \textbf{44.1} \\
 & SmolLM3-3B & \textbf{32.1} & 24.7 & 26.1 \\
\midrule
\multirow{4}{*}{CCI} & Llama-3.2-3B & \textbf{53.0} & 38.6 & 40.7 \\
 & Phi-4-mini & \textbf{40.2} & 27.3 & 28.7 \\
 & Qwen3-4B & \textbf{75.1} & 71.5 & 68.9 \\
 & SmolLM3-3B & \textbf{52.3} & 43.8 & 41.6 \\
\bottomrule
\end{tabular}
\caption{FAPM domain results at sparsity 0.9. The domain-only comparison shows that pruning-based recovery often sacrifices internalized document knowledge, even when it is useful for general-capability restoration.}
\label{tab:fapm}
\end{table*}

\begin{table*}[!t]
\centering
\footnotesize
\begin{tabular}{lllrrrr}
\toprule
Dataset & Model & Variant & Domain (\%) & IFEval (\%) & MMLU (\%) & MSBench (\%) \\
\midrule
\multirow{8}{*}{CC} & \multirow{2}{*}{Llama-3.2-3B} & Vanilla-FAPM & 22.3 & \textbf{75.3} & 51.8 & \textbf{62.0} \\
 & & IA-FAPM & \textbf{22.4} & 74.5 & \textbf{53.7} & 59.0 \\
 & \multirow{2}{*}{Phi-4-mini} & Vanilla-FAPM & \textbf{17.1} & \textbf{56.4} & 58.2 & \textbf{64.5} \\
 & & IA-FAPM & 16.9 & 53.8 & \textbf{61.3} & 63.0 \\
 & \multirow{2}{*}{Qwen3-4B} & Vanilla-FAPM & 41.2 & \textbf{83.8} & 61.3 & 83.5 \\
 & & IA-FAPM & \textbf{44.1} & 82.6 & \textbf{65.5} & \textbf{85.0} \\
 & \multirow{2}{*}{SmolLM3-3B} & Vanilla-FAPM & 24.7 & \textbf{71.0} & 35.5 & \textbf{68.5} \\
 & & IA-FAPM & \textbf{26.1} & 69.4 & \textbf{49.0} & 67.0 \\
\midrule
\multirow{8}{*}{CCI} & \multirow{2}{*}{Llama-3.2-3B} & Vanilla-FAPM & 38.6 & \textbf{75.4} & \textbf{54.2} & \textbf{62.5} \\
 & & IA-FAPM & \textbf{40.7} & 74.7 & 53.7 & 59.5 \\
 & \multirow{2}{*}{Phi-4-mini} & Vanilla-FAPM & 27.3 & 53.6 & 60.0 & \textbf{71.0} \\
 & & IA-FAPM & \textbf{28.7} & \textbf{56.7} & \textbf{60.5} & 64.5 \\
 & \multirow{2}{*}{Qwen3-4B} & Vanilla-FAPM & \textbf{71.5} & 82.7 & 60.0 & \textbf{86.0} \\
 & & IA-FAPM & 68.9 & \textbf{83.7} & \textbf{65.8} & 84.5 \\
 & \multirow{2}{*}{SmolLM3-3B} & Vanilla-FAPM & \textbf{43.8} & \textbf{72.8} & 33.7 & 67.5 \\
 & & IA-FAPM & 41.6 & 72.7 & \textbf{48.2} & \textbf{68.5} \\
\bottomrule
\end{tabular}
\caption{FAPM domain-general results at sparsity 0.9. The table exposes the recovery trade-off directly: FAPM often improves general benchmarks, but the domain column explains why it is not a drop-in replacement for \method{}.}
\label{tab:fapm-full}
\end{table*}

\paragraph{Full pre-recovery recipe grid.}
Tables~\ref{tab:recipe-cc} and~\ref{tab:recipe-cci} report the full pre-recovery Inject recipe grids. On CC, Mixed 1:1:2 is best for Llama and Phi, while Mixed 1:1:1 is best for Qwen and SmolLM. On CCI, Mixed 1:1:2 is best for Llama, Phi, and SmolLM, while reconstruction-only 1:0:0 is best for Qwen.

\begin{table*}[!t]
\centering
\footnotesize
\setlength{\tabcolsep}{3pt}
\begin{tabular*}{\textwidth}{@{\extracolsep{\fill}}lrrrrrr@{}}
\toprule
Model & Vanilla & Mix 1:1:1 & Continue & 1:0:0 & Rewrite & Mix 1:1:2 \\
\midrule
Llama-3.2-3B & 35.5 & 36.1 & 35.1 & 36.8 & 34.4 & \textbf{38.3} \\
Phi-4-mini & 24.4 & 27.5 & 27.2 & 29.9 & 27.6 & \textbf{32.1} \\
Qwen3-4B & 42.4 & \textbf{47.7} & 40.9 & 44.8 & 43.5 & 46.9 \\
SmolLM3-3B & 32.1 & \textbf{36.8} & 32.7 & 33.6 & 31.6 & 36.1 \\
\bottomrule
\end{tabular*}
\caption{CC Inject recipe sensitivity before Recover. Mixed objectives are usually robust on CC, but the best pre-recovery domain checkpoint still varies by model family.}
\label{tab:recipe-cc}
\end{table*}

\begin{table*}[!t]
\centering
\footnotesize
\setlength{\tabcolsep}{3pt}
\begin{tabular*}{\textwidth}{@{\extracolsep{\fill}}lrrrrrr@{}}
\toprule
Model & Vanilla & Mix 1:1:1 & Continue & 1:0:0 & Rewrite & Mix 1:1:2 \\
\midrule
Llama-3.2-3B & 53.0 & 57.2 & 55.8 & 57.2 & 54.4 & \textbf{58.6} \\
Phi-4-mini & 40.2 & 45.0 & 42.1 & 36.7 & 37.7 & \textbf{46.3} \\
Qwen3-4B & 75.1 & 73.7 & 72.9 & \textbf{75.5} & 74.4 & 72.5 \\
SmolLM3-3B & 52.3 & 54.1 & 52.2 & 53.6 & 51.1 & \textbf{54.6} \\
\bottomrule
\end{tabular*}
\caption{CCI Inject recipe sensitivity before Recover. Mixed 1:1:2 remains strongest for most model families, while Qwen3-4B is the high-base-prior boundary case where the 1:0:0 reconstruction recipe slightly exceeds mixed objectives.}
\label{tab:recipe-cci}
\end{table*}

\paragraph{Full Qwen scaling results.}
Table~\ref{tab:scaling-full} reproduces the Qwen scaling scores in the supplementary record for completeness.

\begin{table*}[!t]
\centering
\footnotesize
\setlength{\tabcolsep}{5pt}
\begin{tabular}{llrrrr}
\toprule
Model & Method & Domain (\%) & IFEval (\%) & MMLU (\%) & MSBench (\%) \\
\midrule
\multirow{4}{*}{Qwen3-8B} & Base Instruct & 38.5 & \textbf{87.6} & \textbf{65.3} & \textbf{82.5} \\
 & Vanilla SFT & 48.7 & 56.4 & 14.0 & 52.5 \\
 & Best IA & \textbf{57.5} & 50.6 & 18.5 & 48.5 \\
 & \method{} (TIES $d=0.3$) & \underline{56.8} & \underline{62.2} & \underline{26.7} & \underline{73.5} \\
\midrule
\multirow{4}{*}{Qwen3-14B} & Base Instruct & 40.4 & \textbf{90.0} & \textbf{72.5} & \textbf{81.5} \\
 & Vanilla SFT & 54.8 & 62.9 & 54.5 & 57.0 \\
 & Best IA & \textbf{60.5} & 53.5 & 40.3 & 42.0 \\
 & \method{} (TIES $d=0.3$) & \underline{59.6} & \underline{67.5} & \underline{67.2} & \underline{73.5} \\
\midrule
\multirow{4}{*}{Qwen3-32B} & Base Instruct & 47.2 & \textbf{87.5} & \textbf{74.8} & \textbf{84.5} \\
 & Vanilla SFT & 56.4 & 58.5 & 44.0 & 56.5 \\
 & Best IA & \textbf{63.9} & 53.0 & 63.0 & 44.5 \\
 & \method{} (TIES $d=0.3$) & \underline{62.8} & \underline{67.0} & \underline{74.5} & \underline{72.5} \\
\bottomrule
\end{tabular}
\caption{Complete Qwen3 scaling ablation on CC. Bold and underline mark the best and second-best result within each model block.}
\label{tab:scaling-full}
\end{table*}

\end{document}